# Large-scale benchmarking of multi-objective soft-computing metaheuristics for redundancy allocation in repairable k-out-of-n systems

Mateusz Oszczypała[1], David Ibehej[2], and Jakub Kůdela[2,*]

[1]Military University of Technology, gen. Sylwestra Kaliskiego Street 2, 00-908 Warsaw, Poland.

[2]Institute of Automation and Computer Science, Brno University of Technology, Brno, Czech Republic.

[*] Corresponding author: Jakub.Kudela@vutbr.cz

**Abstract:** This paper presents a large-scale budget-aware benchmark of multi-objective soft-computing metaheuristics for a bi-objective redundancy allocation problem in repairable k-out-of-n systems. The problem combines cost minimization and steady-state availability maximization under weight constraints, with binary subsystem-level decisions determining both the number of redundant components and the redundancy strategy. Four strategies are considered: cold standby, warm standby, hot standby, and a mixed active–warm standby strategy. Subsystem availability is evaluated using continuous-time Markov chains, while the optimization task is treated as a constrained mixed-integer multi-objective problem. The study compares 65 metaheuristic algorithms from multiple algorithmic families under two initialization settings, with and without Scaled Binomial Initialization (SBI), across six case studies of increasing structural and dimensional complexity and four weight limits per case. Performance is assessed using hypervolume, budget-dependent convergence profiles, and non-parametric statistical comparisons. The results show that algorithm rankings are strongly budget-dependent, indicating that a single final-budget ranking can be misleading. SBI provides a substantial early advantage and can change the relative performance of competing methods, especially for larger instances. The best-performing algorithms vary across budget regimes: NNIA-SBI and CMOPSO-SBI are competitive under tight budgets, whereas NSGA-II+ARSBX-SBI performs robustly for medium and large budgets. From a system design perspective, Pareto-optimal solutions are dominated by hot standby and mixed redundancy strategies, while cold and warm standby rarely appear. The benchmark highlights the importance of initialization, computational budget, and problem complexity when selecting soft-computing optimizers for practical redundancy allocation.

## 1. Introduction

Ensuring long-term operation of a system without unplanned downtimes caused by failures is one of the key challenges faced by engineers across various industrial sectors, such as manufacturing [1,2], energy [3,4], transportation [5], and the space industry [6]. In non-repairable systems, where a failure is equivalent to the end of system operation with no return to a functional state, the primary objective is to maximize the probability of correct system performance from the start of operation until the end of its intended operational lifetime [7,8]. In contrast, repairable systems have the ability to return to a functional state through a renewal process following a failure. In this case, besides preventing failures of the system and its components, a crucial aspect is also the ability to restore functionality as quickly as possible in order to maximize system availability [9,10].

The primary method for increasing system reliability and availability is structural redundancy, which involves allocating additional components to subsystems to replace primary elements in the event of a failure [11–14]. The failure rates and the activation method of the additional components determine the redundancy strategy [15,16]. The simplest form is active redundancy, in which redundant components operate simultaneously with the primary ones, ensuring uninterrupted system operation after any single component failure; however, this approach entails high operational and maintenance costs. Standby strategies are classified as cold, warm, or hot, depending on the failure characteristics during the standby state and the switching time from standby to active mode [17,18]. The most complex—and often the most effective—are mixed strategies, which combine multiple basic redundancy modes to minimize the impact of primary component failures while avoiding excessive costs of system construction, maintenance, and renewal [19,20].

Because of the complex structure of the redundancy allocation problem (RAP), metaheuristic algorithms are the primary methods that are currently used to approach it. These methods are most suitable for difficult problems without a known exploitable structure [21] in a wide range of applications such as thermal engineering [22], carbon capture [23] or UAV path planning [24]. Prior RAP research has predominantly examined metaheuristics based on evolutionary computation and swarm intelligence, while largely excluding derivative-based optimization. In this setting, gradient-driven methods are generally of limited applicability due to strongly non-linear and tightly coupled model structures and the frequent presence of discrete or mixed-integer decision variables, which violate the regularity assumptions required for robust gradient exploitation. Despite the substantial volume of published work, empirical comparisons remain disproportionately confined to a narrow set of widely used metaheuristics, thereby restricting a systematic and representative assessment of performance across the broader landscape of established algorithms.

As most metaheuristic methods are hard to analyze analytically, their utility is usually assessed through benchmarking [25,26]. Recent theoretical works for multi-objective methods are mostly focused on the runtime analysis of established methods (like NSGA-II) on well-understood discrete problems, such as OneMinMax [27–29] or continuous problems with spherical functions [30]. For multi-objective optimization methods, there are few established benchmark suites, such as the DTLZ [31] and WFG [32]. However, these suits have probably been overused in the last years to the point where one can speculate on overfitting of optimization algorithms to these problems [33]. The current efforts in benchmarking multi-objective algorithms are on the development and analysis of new suits, such as the ones in the COCO Platform [34–36]. Another current line of research is in bringing benchmarking closer to practical applications by focusing on real-world inspired benchmarks [37,38]. However, although there is an abundance of real-world multi-objective optimization problems, there are only a handful of problems/suits that are readily available for researchers (with code and results publicly available), such as the Radar Waveform problem [39], the HBV Benchmark Problem [40] or the Mazda Benchmark Problem [41].

In this paper, we present a large-scale benchmarking study for a bi-objective redundancy allocation problem in repairable k-out-of-n: G systems. The considered problem provides a reproducible real-world benchmark setting for constrained mixed-integer multi-objective soft-computing metaheuristics, combining discrete subsystem-level design decisions, nonlinear CTMC-based availability evaluation, explicit weight constraints, multiple redundancy strategies, initialization effects, and budget-dependent performance analysis. The test problems are derived from an extensive literature survey and cover various system configurations and structural characteristics, as schematically illustrated in Fig. 1. The developed models assume heterogeneous components at the system level and homogeneous components within each subsystem, with time-invariant failure and repair rates. The CTMC models distinguish operational, standby, and failed component states associated with four redundancy strategies: cold standby, warm standby, hot standby, and a mixed active–warm standby strategy. Overall, 65 metaheuristic algorithms are evaluated under two initialization settings and large computational budgets. These elements define the contribution of this work as a reproducible, budget-aware benchmark for multi-objective soft-computing metaheuristics applied to a constrained repairable-system design problem.

Accordingly, this study addresses the following research questions:

- **RQ1**: How do different redundancy strategies appear in Pareto-optimal designs under varying weight constraints and system complexity?

- **RQ2:** How does the computational budget affect the relative performance of multi-objective soft-computing metaheuristics for the considered RAP?
- **RQ3:** To what extent does Scaled Binomial Initialization influence convergence speed, hypervolume performance, and algorithm ranking?
- **RQ4:** How does case-study complexity, expressed by system structure and problem dimensionality, affect convergence behavior and the computational effort required to obtain high-quality Pareto fronts?

The remainder of this paper is organized as follows. Section 2 reviews recent RAP literature and identifies the main research gaps. Section 3 presents the problem formulation and CTMC-based availability models for the considered redundancy strategies. Section 4 defines the benchmark design, including the numerical case studies, solution encodings, initialization schemes, algorithm portfolio, and experimental setup. Section 5 reports and discusses the optimization results, including redundancy-strategy composition, hypervolume convergence, budget-dependent statistical rankings, relative-distance analysis to the best-found hypervolume, algorithm-family-level performance patterns, and the effect of case-study complexity. Section 6 concludes the paper and outlines future research directions.

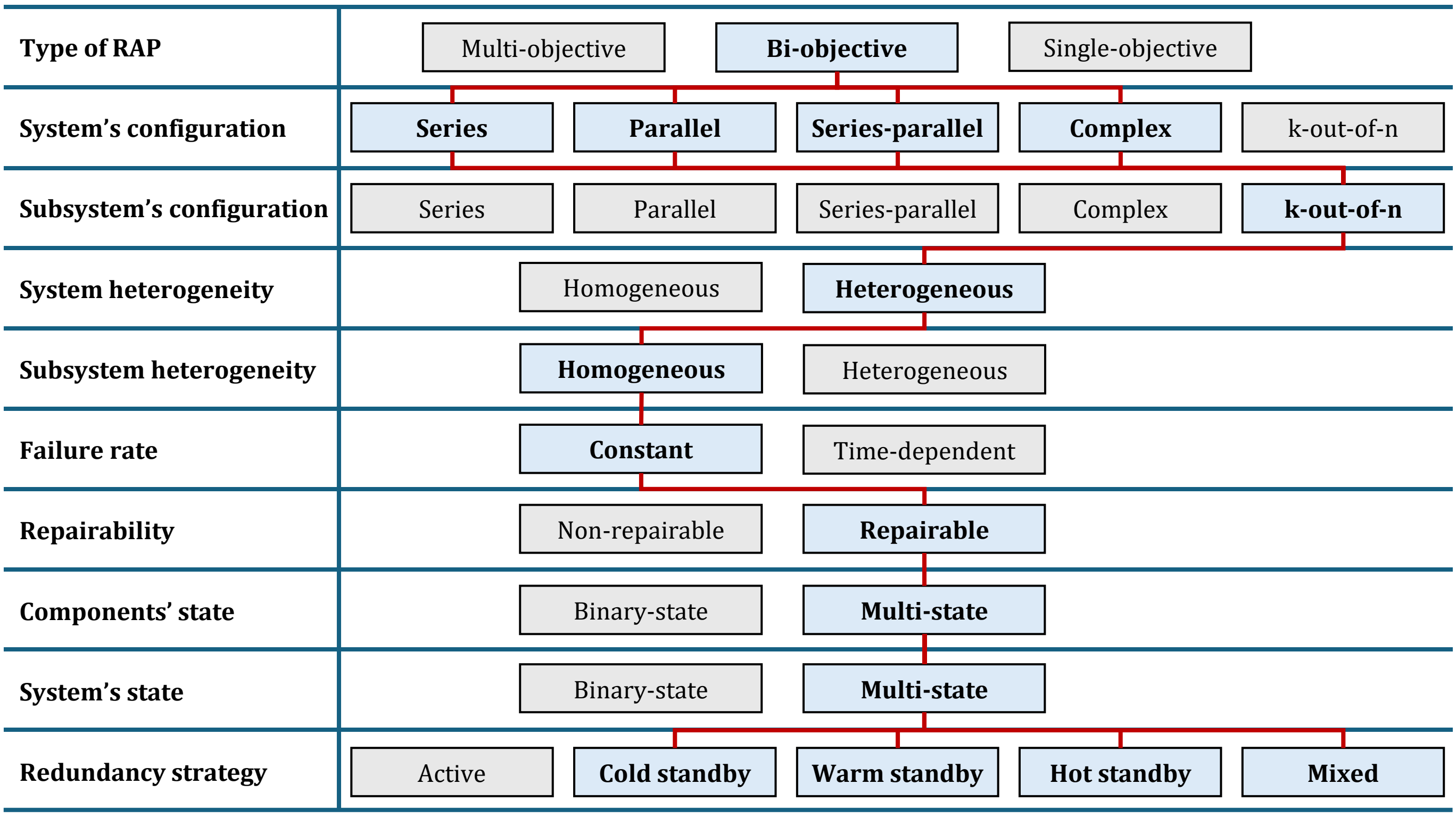

**Fig. 1.** Taxonomy of system configurations and characteristics.

**Notations**

| | |
|---|---|
| $\lambda_{\text{working}}$ | Failure rate of active components in system working states |
| $\lambda_{\text{standby}}$ | Failure rate of warm standby |
| $\sigma_{\text{cold}}$ | Switch rate of component from cold standby to active state |
| $\sigma_{\text{warm}}$ | Switch rate of component from warm standby to active state |
| $\mu$ | Repair rate of failed component |
| $m$ | Number of subsystems |
| $\boldsymbol{k}$ | Minimum working component numbers set for subsystems, $\boldsymbol{k} = \{k_1, k_2, \dots, k_m\}$ |
| $\boldsymbol{n}$ | Component numbers set, $\boldsymbol{n} = \{n_1, n_2, \dots, n_m\}$ |
| $\boldsymbol{s}$ | Standby component numbers set for subsystems, $\boldsymbol{s} = \{s_1, s_2, \dots, s_m\}$ |
| $\boldsymbol{c}$ | Unit costs of components, $\boldsymbol{c} = \{c_1, c_2, \dots, c_m\}$ |
| $\boldsymbol{w}$ | Component weight set, $\boldsymbol{w} = \{w_1, w_2, \dots, w_m\}$ |
| $C_S$ | System cost |
| $W$ | Weight constraint |
| $A_i$ | Availability of $i$-th subsystem |
| $U_i$ | Unavailability of $i$-th subsystem |
| $A_S$ | System availability |
| $\boldsymbol{\Lambda}_S$ | Transition matrix of a CTMC |
| $\boldsymbol{S}$ | State space of a CTMC |
| $\boldsymbol{\Pi}$ | Ergodic probabilities of a CTMC |

## 2. Related works

In recent years, bi-objective RAPs have increasingly attracted scholarly attention due to their practical significance in enhancing system performance under multiple competing objectives. Table 1 provides a summary of recent studies published between 2021 and 2025. Predominantly, existing research efforts have sought to maximize system reliability or availability while simultaneously minimizing total implementation and operational costs. In the context of non-repairable systems, optimization is primarily directed toward the maximization of the reliability function over a specified time horizon [42,43], or alternatively, toward maximizing the Mean Time to First Failure (MTTFF) [44,45]. Conversely, for repairable systems, system availability emerges as the principal performance criterion [46]. The extant body of literature is largely concentrated on active redundancy schemes, whereas comparatively fewer contributions consider mixed redundancy strategies involving combinations of active, cold, and warm standby components. Nevertheless, recent studies such as those by Chambari et al. [47], Juybari et al. [48] and Yaghtin et al. [49] have demonstrated the potential of mixed configurations in improving system-level trade-offs between reliability and cost.

From an algorithmic standpoint, the resolution of bi-objective RAPs has predominantly relied on metaheuristic techniques, particularly those rooted in evolutionary computation and swarm intelligence. Among these, the Non-dominated Sorting Genetic Algorithm II (NSGA-II) has emerged as the most widely employed method, owing to its capacity for approximating high-quality Pareto frontiers across diverse problem instances. Several enhancements to this algorithm have been proposed to address the computational and modeling challenges inherent in RAPs under uncertainty. For instance, Chambari et al. [47] introduced a simulation-based NSGA-II framework capable of accurately estimating reliability functions for systems with heterogeneous and mixed redundancies, albeit at the cost of increased computational effort. Similarly, Hadinejad and Amiri [44] presented a simulation-based model incorporating non-exponential failure and repair rates, further expanding the applicability of RAPs to real-world scenarios.

In parallel, a range of alternative algorithms has been proposed. Yeh et al. [43] reported that the Multi-objective Simplified Swarm Optimization (MOSSO) algorithm outperforms both NSGA-II and MOPSO in terms of convergence speed and solution diversity. Li et al. [42] developed the DMOEA-εC algorithm based on the ε-constraint framework, demonstrating superior performance for uncertain stochastic RAPs. Furthermore, various studies have introduced fuzzy and interval-based approaches to address epistemic and aleatory uncertainties, including hesitant fuzzy modeling [50], triangular fuzzy representations [51], and interval-based crowding distance mechanisms [52–54].

While the majority of studies prioritize reliability and cost objectives, a growing subset of research is oriented toward availability-centric formulations, especially for repairable and load-sharing systems [46,49,55]. Zaretalab et al. [46] proposed a multi-objective RAP model (MORAP) that incorporates supplier selection and reliability-enhancing measures, demonstrating improved performance using the Non-dominated Ranked Genetic Algorithm (NRGA). Similarly, the work of [45] highlighted the impact of optimizing switch reliability and repair policies, achieving substantial improvements in MTTFF with manageable increases in cost. Notably, innovations have also been observed in algorithmic design, including the use of advanced initialization schemes [56] such as the Scaled Binomial Initialization (SBI) for NSGA-II, which was shown to enhance performance in large-scale RAP instances [55].

Table 1 shows that recent bi-objective RAP studies have mainly advanced the field along three directions: richer uncertainty modeling, problem-specific algorithmic adaptations, and the inclusion of selected standby or mixed redundancy strategies. However, from a benchmarking perspective, several limitations remain. Most studies compare only a small number of algorithms, usually including NSGA-II, MOPSO, MOEA/D, or their problem-specific variants. Moreover, the reported comparisons are typically based on final outcomes rather than on budget-dependent convergence behavior. As a result, existing studies provide limited evidence on whether the observed algorithmic superiority is robust across computational budgets, initialization schemes, constraint levels, and system sizes.

The existing RAP literature is therefore still fragmented from the perspective of soft-computing benchmarking. Many studies introduce a new model or a modified algorithm and validate it against a small set of baselines. Such comparisons are useful for assessing individual methods, but they are less suitable for identifying general performance patterns across algorithmic families. In particular, limited attention has been paid to out-of-the-box robustness, initialization sensitivity, and the dependence of algorithm rankings on the available evaluation budget. These aspects are important in practical RAP applications, where objective-function evaluations may be computationally expensive and where the best final performer may not be the best choice under limited computational resources.

Based on this review, four gaps can be identified. First, algorithmic comparisons in RAP studies are usually narrow and centered on a small set of established methods. Second, the effect of computational budget is rarely analyzed explicitly, even though different algorithms may dominate at different stages of the search. Third, initialization mechanisms are usually treated as implementation details rather than as factors that can influence convergence and ranking. Fourth, availability-oriented repairable RAPs with multiple standby strategies remain less systematically benchmarked than reliability-oriented or non-repairable formulations.

To the best of our knowledge, no previous study has combined these elements in a single controlled benchmark: repairable k-out-of-n:G systems, CTMC-based availability evaluation, four redundancy strategies, broad multi-objective algorithm coverage, two initialization settings, and budget-dependent performance analysis. These gaps motivate the present study, which treats the repairable RAP not only as a reliability-design problem, but also as a reproducible real-world benchmark for constrained mixed-integer multi-objective soft-computing algorithms.

**Table 1**
A summary of related works on bi-objective RAP from 2021 to 2025 (MR – max reliability, MA – max availability, MC – min cost, A/H – active/hot standby, W – warm standby, C – cold standby, M - mixed).

| Paper | Year | Objective functions | | | Redundancy strategies | | | | Algorithms | Main conclusions |
|---|---|---|---|---|---|---|---|---|---|---|
| | | MR | MA | MC | A/H | W | C | M | | |
| [43] | 2021 | ✓ | × | ✓ | ✓ | × | × | × | MOPSO, MOSSO, NSGA-II | MOSSO algorithm outperforms NSGA-II and MOPSO in convergence and solution diversity for bi-objective active RRAP with mixed variables |
| [42] | 2021 | ✓ | × | ✓ | × | × | ✓ | × | MOEA/D-AWA, DMOEA-εC | DMOEA-εC algorithm outperforms MOEA/D-AWA in solving medium- and large-scale bi-objective uncertain stochastic RAPs by delivering superior Pareto solutions under risk and cost objectives using an ε-constraint framework |
| [47] | 2021 | ✓ | × | ✓ | ✓ | × | ✓ | ✓ | NSGA-II | Simulation-based NSGA-II approach outperformed existing methods in cost and reliability, demonstrating the effectiveness of using stochastic simulation for bi-objective RAP |
| [48] | 2022 | ✓ | × | ✓ | ✓ | × | ✓ | ✓ | NSGA-II | Mixed redundancy strategy using a heterogeneous backup scheme and CTMC modeling outperforms traditional approaches by enhancing system reliability and cost-efficiency through optimal sequencing of standby components |
| [46] | 2022 | × | ✓ | ✓ | ✓ | × | × | × | NSGA-II, NSGA-III, NRGA, MOEA/D | MORAP model with supplier selection and reliability-enhancing activities showed high applicability, with NRGA outperforming NSGA-II in solution quality |
| [50] | 2022 | ✓ | × | ✓ | ✓ | × | × | × | MOPSO | Hesitant fuzzy MOPSO algorithm effectively handles uncertainty and achieves enhanced reliability and cost efficiency in system design under hesitant environments |
| [54] | 2023 | ✓ | × | ✓ | ✓ | × | × | × | IP-ICA-MOEA, IP-MOEA, IC-MOEA, MI-MOEA, Mid-NSGA-II, | IP-ICA-MOEA with angle-based interval crowding distance significantly improves performance and reduces runtime in bi-objective RAP under uncertainty |
| [53] | 2023 | ✓ | × | ✓ | ✓ | × | × | × | TNF-MOPSO | TNF-MOPSO algorithm outperforms MOPSO and NF-MOPSO in solving time-based fuzzy MORRAP by delivering more stable solutions across reliability, cost, and repair cost objectives |
| [51] | 2023 | ✓ | × | ✓ | ✓ | × | × | × | NSGA-II, NF-MOPSO | TF-MORRAP model effectively captures time-based uncertainty using triangular fuzzy numbers, with NF-MOPSO outperforming NSGA-II in reliability and cost over time |
| [45] | 2024 | ✓ | × | ✓ | × | ✓ | × | × | NSGA-II | NSGA-II-based approach demonstrates that optimizing switch reliability and repair policies can drastically enhance system MTTF, achieving up to 40-fold improvements at moderate cost increases, thus enabling informed trade-offs between performance and expense in RAP design |
| [52] | 2024 | ✓ | × | ✓ | ✓ | × | × | × | NSGA-II, NSGA-II-CDE, NSGA-II-AGDV | NSGA-II-AGDV algorithm effectively solves TIVF-MORRAP by outperforming existing methods in reliability, cost, and repair cost under time- and interval-based uncertainty |
| [44] | 2024 | ✓ | × | ✓ | ✓ | × | ✓ | × | OVS | Simulation-based multi-objective optimization model integrates non-exponential failure and repair rates with practical system constraints, delivering valid and efficient solutions to RAPs by maximizing MTTFF and minimizing total cost under real-world conditions |
| [49] | 2025 | × | ✓ | ✓ | ✓ | × | ✓ | ✓ | NSGA-II | NSGA-II-based approach effectively models load-sharing systems under shock attacks, delivering balanced Pareto-optimal solutions that enhance system reliability while minimizing costs in complex series-parallel configurations |
| [55] | 2025 | × | ✓ | ✓ | × | × | × | ✓ | NSGA-II | SBI initialization method for NSGA-II demonstrates superior performance compared to standard random initialization, particularly in large-scale systems |
| This paper | 2026 | × | ✓ | ✓ | ✓ | ✓ | ✓ | ✓ | 65 metaheuristics | A large-scale budget-aware benchmark identifying how algorithm performance depends on computational budget, initialization scheme, system complexity, and weight constraints, while analyzing redundancy-strategy patterns in Pareto-optimal solutions. |

CTMC – Continuous- Time Markov Chain.
DMOEA-$\varepsilon$C – Decomposition-Based Multi-Objective Evolutionary Algorithm with the $\varepsilon$-Constraint Framework.
IP-ICA-MOEA – Imprecision-Propagating Interval Crowding Distance Multi-objective Optimization Algorithm.
IP-MOEA – Imprecision-Propagating Multi-objective Optimization Algorithm.
MOEA/D – Multi-objective Evolutionary Algorithm Based on Decomposition.
MOEA/D-AWA – Multi-objective Evolutionary Algorithms based on Decomposition with Adaptive Weight Adjustment.
MOHNS – Multi-objective Hybrid Metaheuristic.
MOPSO – Multi-objective Particle Swarm Optimization.
MOSSO – Multi-objective Simplified Swarm Optimization.
MTTFF – Mean Time to First Failure.
NRGA – Non-dominated Ranked Genetic Algorithm.
NSGA-II – Non-dominated Sorting Genetic Algorithm-II.
NSGA-II-AGDV – NSGA-II with Agglomerative and Divisive Clustering Algorithms.
NSGA-II-CDE – NSGA-II with Crowding Distance Elimination.
NSGA-III – Non-dominated Sorting Genetic Algorithm-III.
OVS – Optimization Via Simulation.
SBI – Scaled Binomial Initialization.
TNF-MOPSO – Tuning and Neighborhood based Fuzzy Multi-Objective Particle Swarm Optimization.

## 3. Methodology

The proposed modeling framework is based on the following assumptions regarding the system structure and availability evaluation:

1) Each subsystem is modeled as a k-out-of-n:G structure, in which at least k out of n components must be operational for the subsystem to function correctly [53];
2) Four redundancy strategies are considered: cold standby, warm standby, hot standby, and a mixed active–warm standby strategy. The mixed strategy is defined as a specific configuration in which one redundant component operates in active mode, while the remaining redundant components are maintained in warm standby;
3) The system is composed of heterogeneous subsystems, whereas components within each subsystem are assumed to be homogeneous;
4) Components are represented as being in one of three operational states: active, standby, or failed. These states describe the operational role and failure status of a component, rather than distinct performance levels in the classical multi-state reliability framework;
5) Failed components are assumed to be repairable and return to the standby state after repair completion;
6) Failure, repair, and standby-to-active transition times are assumed to be exponentially distributed, which enables continuous-time Markov chain-based availability modeling;
7) Failure, repair, and switching parameters are modeled as transition rates expressed per unit time, rather than as probabilities. Consequently, the corresponding numerical values may be greater than one depending on the adopted time scale;
8) Redundancy strategies are assumed to remain fixed over the considered operating period. Time-dependent, adaptive, or dynamically reconfigured redundancy policies are not considered in the present study;
9) Component transitions are assumed to be statistically independent, and common-cause failures are not considered.

### *3.1. Problem formulation*

In repairable systems, the primary objective is to ensure high availability levels while concurrently minimizing system acquisition and deployment costs. The analyzed bi-objective RAP formulation seeks to simultaneously maximize system availability and minimize total cost, under a predefined weight constraint.

Problem definition:

$$\text{Maximize } A_{\boldsymbol{S}}(\boldsymbol{n},\boldsymbol{k},\boldsymbol{r}) = f(A_1(n_1,k_1,r_1),\dots,A_i(n_i,k_i,r_i),\dots,A_1(n_m,k_m,r_m)) \tag{1}$$

$$\text{Minimize } C_{\boldsymbol{S}}(\boldsymbol{n}) = \sum_{i=1}^{m} c_i n_i \tag{2}$$

Subject to:

$$\sum_{i=1}^{m} w_i n_i \leq W \tag{3}$$

where $n_i$ and $k_i$ are natural numbers satisfying the condition:

$$n_i \geq k_i, i = 1, 2, \ldots, m. \tag{4}$$

From a probabilistic perspective, the availability of the $i$-th subsystem, configured with a k-out-of-n: G structure, is quantified as the cumulative probability that the subsystem resides in states where at least $k$ components are functioning. This can be formally represented as:

$$A_i(n_i, k_i, r_i) = \sum_{j \in S_{\text{working}}} \pi_j, \tag{5}$$

where $S_{\text{working}}$ denotes the set of operational states in which the subsystem meets or exceeds the minimum operational threshold $k$. The size of this state set is determined by the subsystem parameters $n_i$, $k_i$, and the redundancy strategy $r_i$. The steady-state probabilities $\pi_i$ are obtained by solving the balance equations of the CTMC, given by:

$$\mathbf{\Pi} \cdot \mathbf{\Lambda_S} = 0, \tag{6}$$

subject to the normalization constraint:

$$\sum_{i \in S} \pi_i = 1. \tag{7}$$

### *3.2. Modified continuous-time Markov chains (CTMC) for availability evaluation*

To evaluate subsystem availability in repairable k-out-of-n: G structures, continuous-time Markov chain (CTMC) models are constructed for each redundancy strategy considered in this study. The models provide a consistent strategy-aware availability-evaluation procedure for candidate RAP solutions generated by the optimization algorithms. For each strategy, the CTMC state space distinguishes the numbers of active, standby, and failed components, and steady-state availability is computed as the cumulative probability of the operational states satisfying the k-out-of-n: G requirement. In contrast to previous studies [53,55,57], this work focuses on the development of generalized, strategy-aware models and their software-level implementation for generic k-out-of-n configurations. This modeling framework enables broad applicability designed to support different subsystem sizes and redundancy strategies within a unified computational procedure, thereby enhancing scalability and flexibility in reliability analysis. The developed models address the following redundancy strategies:

- cold standby,
- warm standby,
- mixed strategy (active and warm standby),
- hot standby.

The switching parameters used in the CTMC models represent transition rates, not switching probabilities. Therefore, their values are expressed per unit time and may be greater than one, depending on the adopted time scale. A higher switching rate corresponds to a shorter expected activation time of a standby component.

#### *3.2.1. Cold standby strategy*

Under the adopted ideal cold-standby assumption, standby components are not exposed to failure before activation. The cold standby strategy is based on the assumption that standby components are assumed not to fail before activation. This reflects scenarios in which inactive units are stored under controlled maintenance conditions that minimize their failure likelihood. As a result, this approach contributes to a reduction in system repair costs. However, a key drawback of the strategy is the relatively long activation time required to transition cold standby components into operational mode [58,59].

The system state is described by a three-dimensional vector indicating the number of active components, the number of components in cold standby mode, and the number of failed components. The CTMC model for a k-out-of-n: G system employing the cold standby strategy includes a state space consisting of $n - k + 1$ operational states and $n - k + 2$ failure states. State transitions are driven by three stochastic processes: the failure of an active component, the activation (switch-over) of a component from cold standby mode, and the repair of a failed component. The state transition diagram of CTMC for cold standby strategy, including transition rates, is presented in Fig. 2.

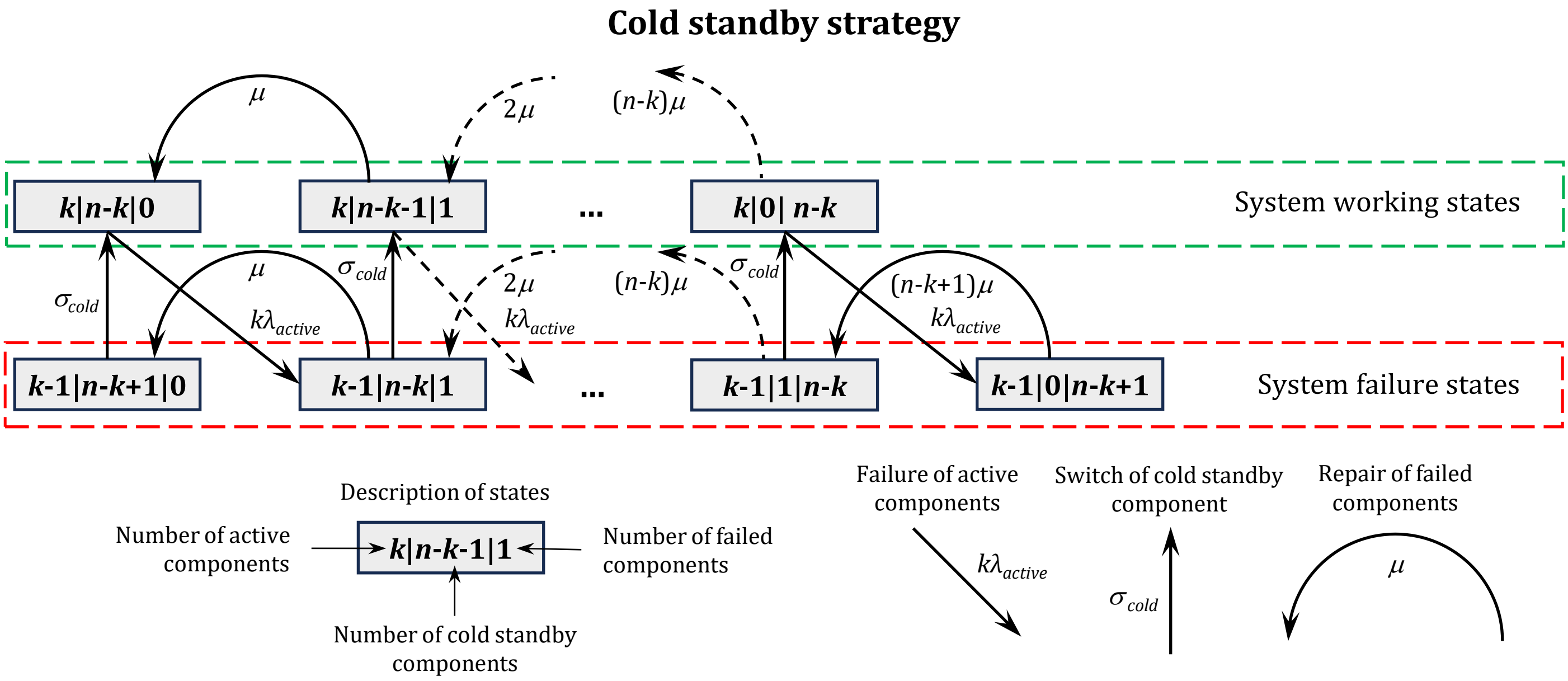

**Fig. 2.** CTMC for cold standby strategy (strategy 0).

#### *3.2.2. Warm standby strategy*

The warm standby strategy assumes that a standby component may fail prior to its activation, which distinguishes it from the cold standby strategy. The failure rate of a component in warm standby is lower than that of an active component, satisfying the condition $\lambda_{\text{active}} > \lambda_{\text{standby}} > 0$ . Nevertheless, a notable advantage of the warm standby approach is the relatively fast activation of standby units, such that $\sigma_{\text{warm}} > \sigma_{\text{cold}}$, which positively influences the availability of subsystems and the entire system [60,61]. Both the warm-standby failure parameter and the standby-to-active switching parameter are modeled as transition rates per unit time.

The CTMC state space for a k-out-of-n: G system employing the warm standby strategy consists of $n - k + 1$ operational states and $n - k + 2$ failure states, identical to the structure observed in the cold standby model. A key distinction between the two models lies in the inclusion of inter-state transitions resulting from failures of warm standby components. A failure of a warm standby unit triggers a transition between two operational states that differ by one fewer standby component and one additional failed component. The CTMC state transition diagram for the warm standby strategy, including the associated transition rates, is presented in Fig. 3.

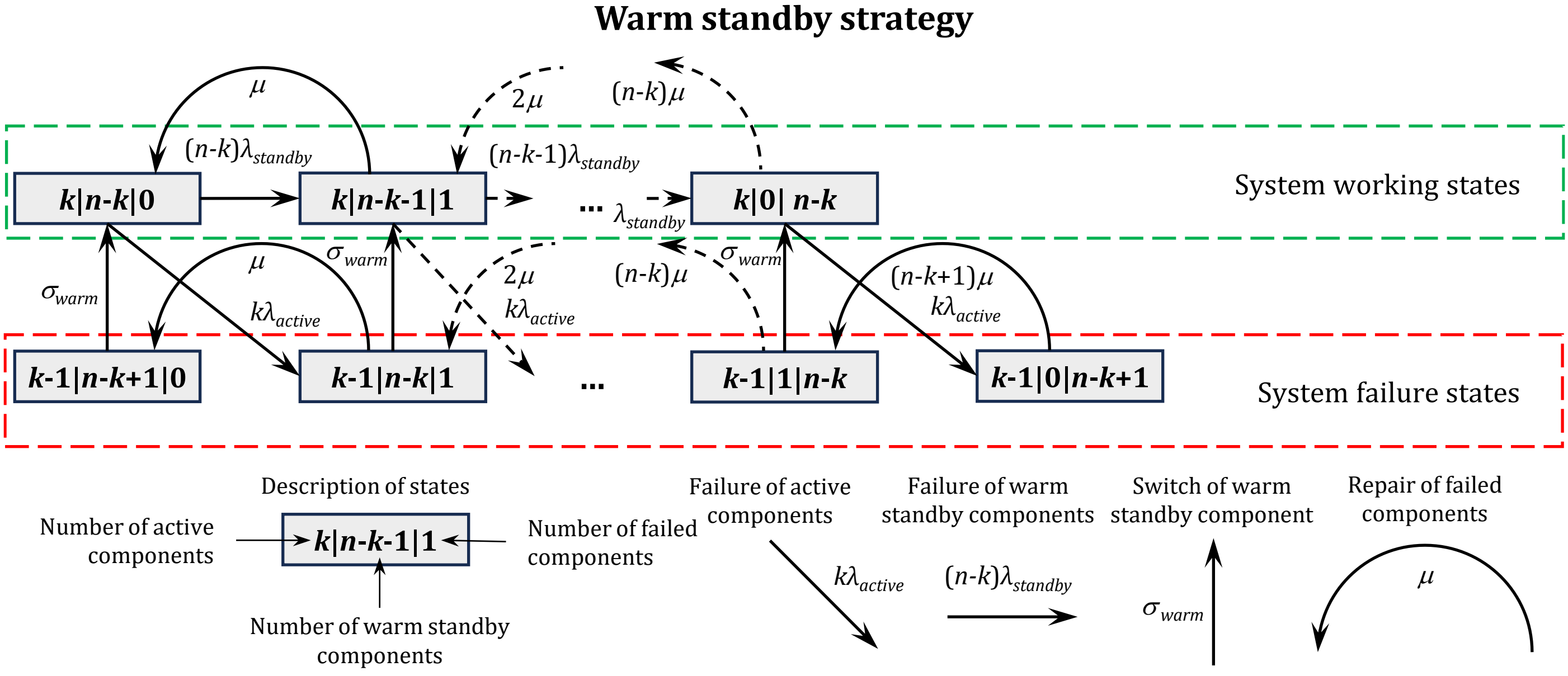

**Fig. 3.** CTMC for warm standby strategy (strategy 1).

### *3.2.3. Mixed strategy (active and warm standby)*

The mixed strategy considered in this study is a specific active–warm standby configuration. It consists of one active redundant component operating under the same failure rate as the primary active components, while the remaining redundant components are maintained in warm standby. Thus, the term mixed strategy does not refer to the full class of possible mixed redundancy policies, but to the active–warm configuration adopted consistently in the benchmark.

The system may operate with $k + 1$, $k$, or $k - 1$ active components. The most desirable condition occurs when $k + 1$ components are active, providing a built-in redundancy that allows the system to tolerate a single component failure without transitioning into a degraded state. Within the correct operational regime of the k-out-of-n system, there are $n - k$ states with $k + 1$ active components and $n - k + 1$ states with exactly $k$ active components. Conversely, system failure is represented by $n - k + 2$ states in which only $k - 1$ components are active. Consequently, the complete state space of the system consists of $3n - 3k + 3$ states, of which $2n - 2k + 1$ correspond to proper system operation.

The desired operational state is characterized by $k + 1$ components functioning in active mode, with the remaining $n - k - 1$ units operating in warm standby, and no component in a failed condition. Active components are subject to failure at a rate of $\lambda_{\text{active}}$, while those in warm standby fail at a lower rate, $\lambda_{\text{standby}}$. Given the assumption that failures of individual components occur independently, the overall rate of transition from the current state due to failure is the sum of the failure rates of all active or standby units. When the system contains fewer than $k + 1$ active components, but at least one unit remains in warm standby, one or two standby components may be activated as needed. This activation occurs at a rate of $\sigma_{\text{warm}}$ for a single component or $2\sigma_{\text{warm}}$ for a dual activation. Failed units are restored at a repair rate $\mu$, and upon completion of the repair process, they return to the warm standby pool. They are only promoted to active status when operational demands require it. Fig. 4 illustrates the state transition diagram of the CTMC corresponding to the mixed redundancy strategy, along with the respective transition rates.

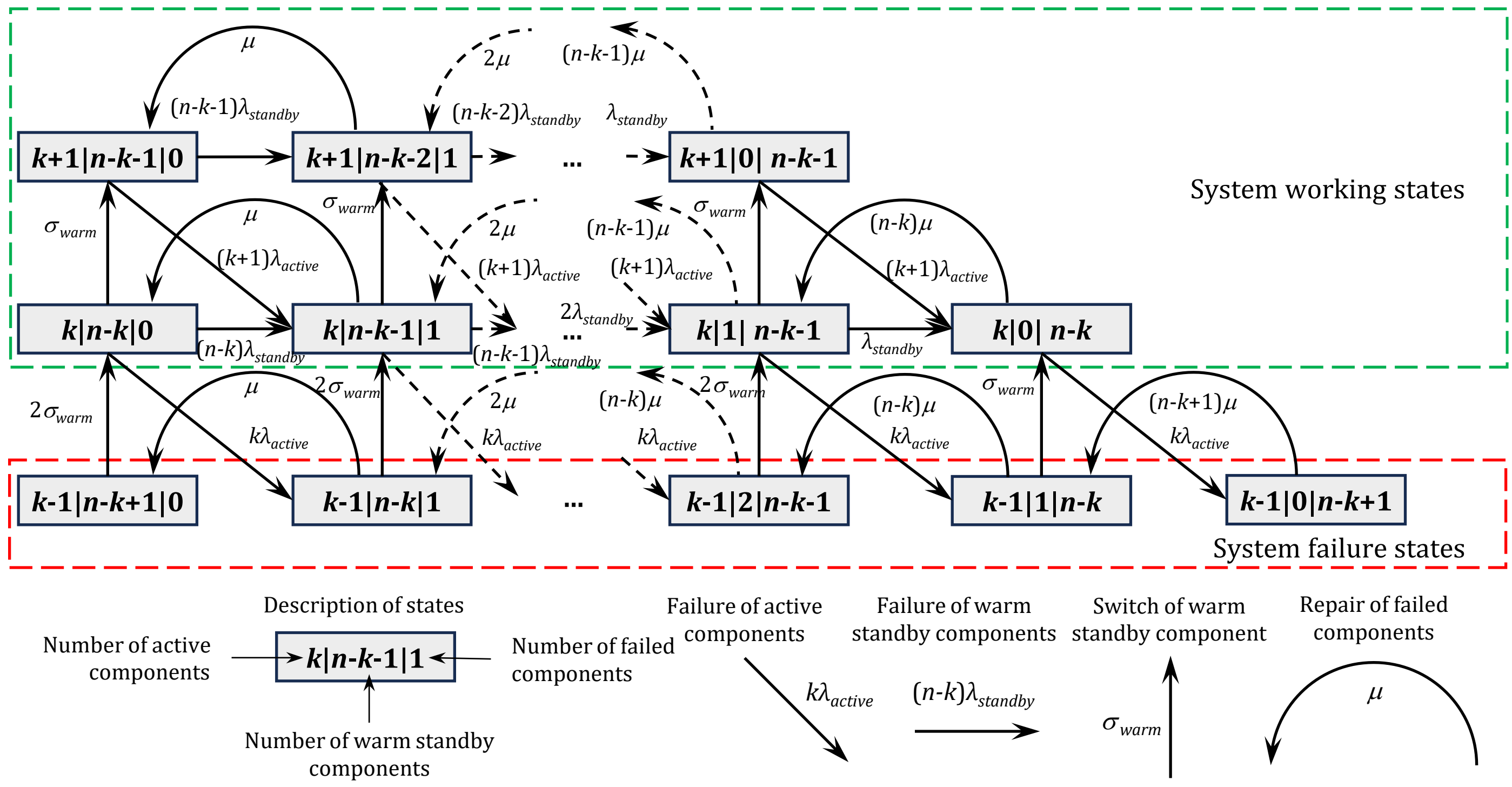

**Fig. 4.** CTMC for mixed strategy (strategy 2).

### 3.2.4. *Hot standby strategy*

The hot standby strategy assumes immediate substitution of a failed active component by an already active redundant unit [62,63]. As a result, the CTMC model for this strategy comprises $n-k+1$ distinct states, of which $n-k$ represent operational conditions where system functionality is retained. System failure is represented by a single down state, in which only $k-1$ components remain operational and $n-k+1$ have failed. Transitions to states with a reduced number of functioning components are driven by component failures and occur at a rate proportional to the total number of active and standby units, multiplied by the failure rate of active components. In contrast, transitions to states with a higher number of functioning components arise from repairs, occurring at a rate equal to the number of failed components times the repair rate. Perfect switching from standby to active mode, along with identical failure rates for active and standby components, are characteristic features of the hot standby and active redundancy strategies in many industrial systems. A graphical representation of the CTMC for the hot standby strategy, including all associated transition rates, is provided in Fig. 5.

Hot standby strategy

System working states

System failure state

$\mu$ $2\mu$ $(n\text{-}k)\mu$ $(n\text{-}k+1)\mu$

k|n-k|0 → k|n-k-1|1 → ... → k|0| n-k → k-1|0|n-k+1

$n\lambda_{active}$ $(n\text{-}1)\lambda_{active}$ $(k+1)\lambda_{active}$ $k\lambda_{active}$

Description of states

Number of active components — k|n-k-1|1 — Number of failed components

Number of hot standby components

Failure of active and hot standby components $k\lambda_{active}$

Repair of failed components $\mu$

**Fig. 5.** CTMC for hot standby strategy (strategy 3).

### 3.2.5. *Programming implementation*

The developed CTMC models for k-out-of-n: G subsystems employing one of the four redundancy strategies were implemented in the MATLAB programming environment. The corresponding pseudocode is presented in Algorithm 1.

**Algorithm 1:** Pseudocode of creating CTMC for k-out-of-n system to implement in MATLAB

```
1    Identify the redundancy strategy
2    Create the system working states
3        switch (redundancy strategy)
4            case 0:
5                Cold standby ← CTMC with n–k+1 working states and n–k+2 failure states
```

```
6               case 1:
7                   Warm standby ← CTMC with n–k+1 working states and n–k+2 failure states
8               case 2:
9                   Mixed strategy ← CTMC with 2n–2k+1 working states and n–k+2 failure states
10              case 3:
11                  Hot standby ← CTMC with n–k+1 working states and one failure state
12      tot_num_1 ← the total number of system working states with k+1 active components
13      tot_num_2 ← the total number of system working states with k active components
14      tot_num_3 ← the total number of system failure states
15      j ← 0
16      for i from 1 to tot_num_1 do
17          S(i, 1) ← k+1 (number of active components)
18          S(i, 2) ← j (number of standby components)
19          S(i, 3) ← n–k–j–1 (number of failed components)
20          j ← j+1
21      end for
22      j ← 0
23      for i from (tot_num_1+1) to (tot_num_1+tot_num_2) do
24          S(i, 1) ← k (number of active components)
25          S(i, 2) ← j (number of standby components)
26          S(i, 3) ← n–k–j (number of failed components)
27          j ← j+1
28      end for
29      j ← 0
30      for i from (tot_num_1+tot_num_2+1) to (tot_num_1+tot_num_2+tot_num_3) do
31          S(i, 1) ← k–1 (number of active components)
32          S(i, 2) ← j (number of standby components)
33          S(i, 3) ← n–k–j+1 (number of failed components)
34          j ← j+1
35      end for
36      Calculate the transition rates in CTMC
37      for i from 1 to (tot_num_1+tot_num_2+tot_num_3) do
38          for j from 1 to (tot_num_1+tot_num_2+tot_num_3) do
39              if S(i, 1)–S(j, 1) == 1 and S(j, 3)–S(i, 3) == 1 then
40                  Failure of active component: Λ_S(i, j) ← S(i, 1)·λ_active
41              elseif S(i, 2)–S(j, 2) == 1 and S(j, 3)–S(i, 3) == 1 then
42                  if S(i, 1) == k – 1 then
43                      Failure of active and standby component is impossible: Λ_S(i, j) ← 0
44                  elseif
45                      Failure of warm standby component: Λ_S(i, j) ← S(i, 1)·λ_standby
46                  endif
47              elseif S(i, 2)–S(j, 2) == 1 and S(j, 1)–S(i, 1) == 1 then
48                  switch (redundancy strategy)
49                      case 0: cold standby
50                          Switching of cold standby component Λ_S(i, j) ← min(min(S(i, 2), 2), k+1–S(i, 1))·σ_cold
51                      case 1: warm standby
52                          Switching of warm standby component Λ_S(i, j) ← min(min(S(i, 2), 2), k+1–S(i, 1))·σ_warm
53                      case 2: mixed strategy
54                          Switching of warm standby component Λ_S(i, j) ← min(min(S(i, 2), 2), k+1–S(i, 1))·σ_warm
55              elseif S(i, 3)–S(j, 3) == 1 and S(j, 2)–S(i, 2) == 1 then
56                  Repair of failed component: Λ_S(i, j) ← S(i, 3)·μ
57              else
58                  Transition impossible: Λ_S(i, j) ← 0
59              end if
60          end for
61      end for
62      Calculate the diagonals in the transition matrix
63      for i from 1 to (tot_num_1+tot_num_2+tot_num_3) do
64          Λ_S(i, i) ← negative sum of elements in row i
65      end for
66      return CTMC transition matrix Λ_S
```

## 4. Benchmark design and algorithms

### 4.1. Case studies

The study involves the analysis of six case studies with increasing computational complexity. These range from three systems composed of five subsystems connected in series, series-parallel, and complex bridge configurations, to two larger-scale systems consisting of ten subsystems, and finally to a system comprising fifteen subsystems. This approach enables a comprehensive evaluation of the performance of metaheuristic algorithms across problems with varying levels of complexity and search space dimensionality. The system structures for the six case studies are presented in Fig. 6.

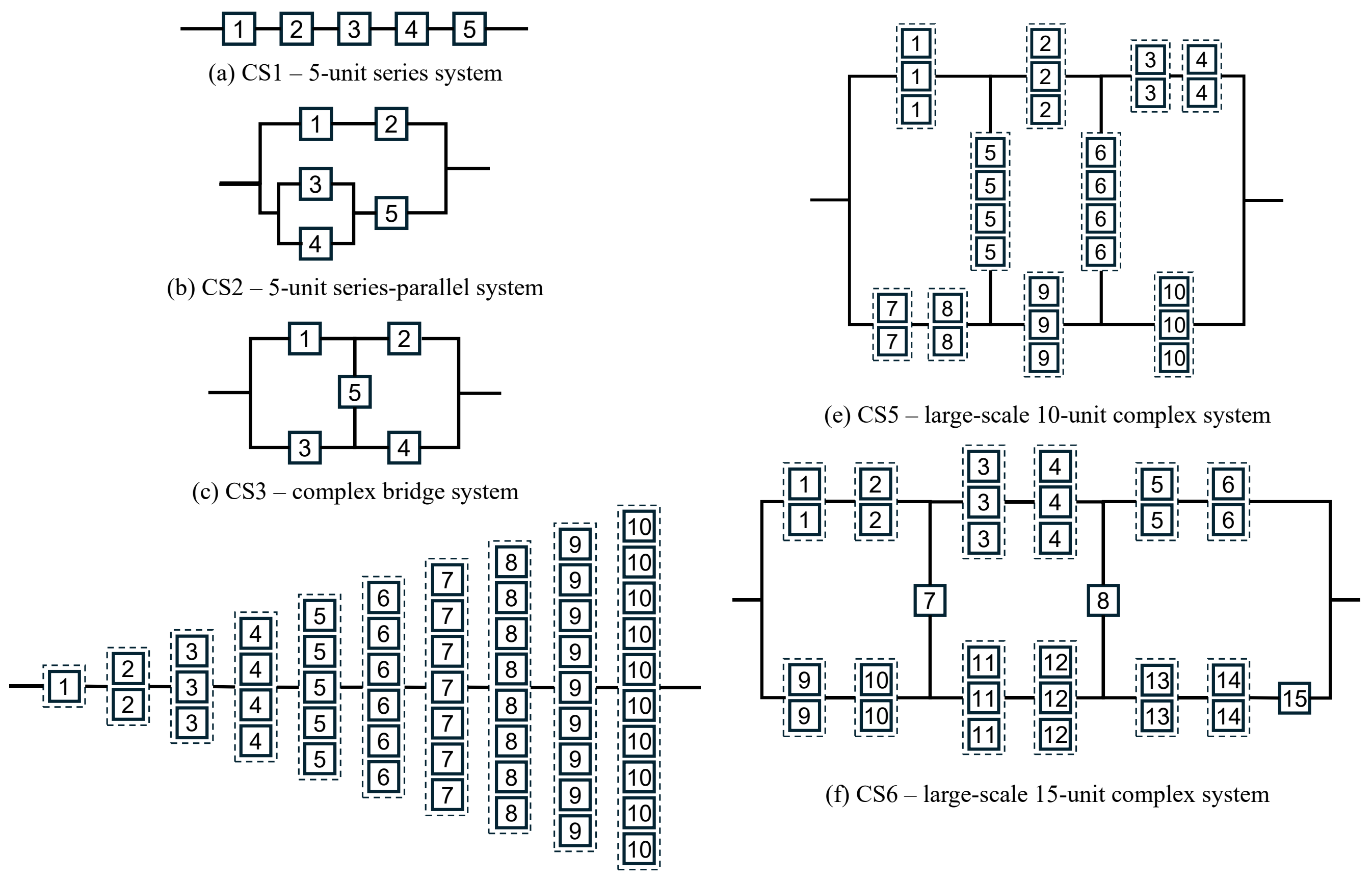

(a) CS1 – 5-unit series system

(b) CS2 – 5-unit series-parallel system

(c) CS3 – complex bridge system

(d) CS4 – large-scale 10-unit series system

(e) CS5 – large-scale 10-unit complex system

(f) CS6 – large-scale 15-unit complex system

**Fig. 6.** System structures of numerical case studies.

System availability is calculated according to the following formulas:

- CS1 – 5-unit series system

$$A_S = \Pi_{i=1}^{5} A_i, \tag{8}$$

- CS2 – 5-unit series-parallel system

$$A_S = 1 - (1 - A_1A_2)(1 - (A_3 + A_4 - A_3A_4)A_5), \tag{9}$$

- CS3 – complex bridge system

$$A_S = A_1A_2 + A_3A_4 + A_1A_4A_5 + A_2A_4A_5 - A_1A_2A_3A_4 - A_1A_2A_3A_5 - A_1A_2A_4A_5 - A_1A_3A_4A_5 - A_2A_3A_4A_5 + 2A_1A_2A_3A_4A_5, \tag{10}$$

- CS4 – large-scale 10-unit series system

$$A_S = \Pi_{i=1}^{10} A_i, \tag{11}$$

- CS5 – large-scale 10-unit complex system

$$\begin{aligned} A_S = {} & A_1A_2A_3A_4 + A_1A_2A_6A_{10}(U_3 + A_3U_4) + A_1A_5A_9A_{10}(U_2 + A_2U_3U_6 + A_2A_3U_4U_6) \\ & + A_7A_8A_9A_{10}(U_1 + A_1U_2U_5 + A_1A_2U_3U_5U_6 + A_1A_2A_3U_5U_6U_4) + A_2A_3A_4A_5A_7A_8U_1(U_9 + A_9U_{10}) \\ & + U_1A_3A_4A_6A_7A_8A_9U_{10}(U_2 + A_2U_5) + A_1U_2A_3A_4A_6A_7A_8A_9U_{10} + A_1U_2A_3A_4A_5A_6A_9U_{10}(U_7+A_7U_8) \\ & + U_1A_2A_5A_6A_7A_8U_9A_{10}(U_3 + A_3U_4), \end{aligned} \tag{12}$$

- CS6 – large-scale 15-unit complex system

$$\begin{aligned} A_S = {} & A_1A_2A_3A_4A_5A_6 + A_9A_{10}A_{11}A_{12}A_{13}A_{14}A_{15}(U_1 + A_1U_2 + A_1A_2U_3 + A_1A_2A_3U_4 + A_1A_2A_3A_4U_5 + A_1A_2A_3A_4A_5U_6) \\ & + A_3A_4A_5A_6A_7A_9A_{10}(U_{11}+A_{11}U_{12} + A_{11}A_{12}U_{13} + A_{11}A_{12}A_{13}U_{14} + A_{11}A_{12}A_{13}A_{14}U_{15}) \cdot (U_1 + A_1U_2) \\ & + ((U_1 + A_1U_2) \cdot (U_3 + A_3U_4 + A_3A_4U_7) + A_1A_2U_7 \cdot (U_3 + A_3U_4)) \cdot (U_{13} + A_{13}U_{14} + A_{13}A_{14}U_{15}) \\ & \cdot A_5A_6A_8A_9A_{10}A_{11}A_{12} + A_1A_2A_5A_6A_7A_8A_{11}A_{12}(A_9A_{10} + U_9 + A_9U_{10}) \cdot (U_3 + A_3U_4) \\ & \cdot (U_{13} + A_{13}U_{14} + A_{13}A_{14}U_{15}) + (U_5 + A_5U_6) \cdot ((U_7 + A_7U_{11} + A_7A_{11}U_{12}) \cdot (U_9+A_9U_{10}) \\ & + A_9A_{10}(U_{11} + A_{11}U_{12}) \cdot A_1A_2A_3A_4A_8A_{13}A_{14}A_{15}) + A_1A_2A_7A_{11}A_{12}A_{13}A_{14}A_{15}(U_9 + A_9U_{10}) \cdot (U_3 + A_3U_4 \\ & + A_3A_4U_5 + A_3A_4A_5U_6) + A_3A_4A_7A_8A_9A_{10}A_{13}A_{14}A_{15}(U_1 + A_1U_2) \cdot (U_{11} + A_{11}U_{12}) \cdot (U_5 + A_5U_6). \end{aligned} \tag{13}$$

Table 2 presents the randomly generated values of technical parameters for the system components and subsystems. The parameter generation process was designed to maintain realistic and meaningful relationships among the variables. In particular, the failure rate of an active component is set higher than that of a warm standby component ($\lambda_{\text{working}} > \lambda_{\text{standby}}$), reflecting the typical operational degradation under load. Additionally, the switching rate of a warm standby component is always greater than that of a cold standby component ($\sigma_{\text{warm}} > \sigma_{\text{cold}}$), consistent with standard standby behavior in reliability modeling. For each case study, four levels of weight constraints were defined to investigate their influence on the performance of the optimization algorithms. These constraints were selected to exceed the total weight of all active components, thereby ensuring system operability and enabling the feasible allocation of redundant components.

**Table 2**
Parameter values for numerical case studies.

| Case study | No of subsystem | $\boldsymbol{k}$ | $\lambda_{\text{working}}$ | $\lambda_{\text{standby}}$ | $\sigma_{\text{cold}}$ | $\sigma_{\text{warm}}$ | $\mu$ | $\boldsymbol{c}$ | $\boldsymbol{w}$ | $W$ |
|---|---|---|---|---|---|---|---|---|---|---|
| CS1, CS2, CS3 | 1 | 1 | 0.75 | 0.50 | 5.30 | 9.99 | 1.14 | 6.50 | 3.56 | {60, 80, 100, 120} |
| | 2 | 1 | 0.26 | 0.17 | 6.65 | 12.54 | 1.87 | 4.30 | 2.87 | |
| | 3 | 1 | 0.51 | 0.34 | 6.08 | 11.46 | 1.58 | 9.42 | 2.05 | |
| | 4 | 1 | 0.70 | 0.47 | 6.99 | 13.18 | 1.55 | 9.67 | 2.54 | |
| | 5 | 1 | 0.89 | 0.59 | 5.16 | 9.73 | 1.14 | 6.95 | 2.21 | |
| CS4 | 1 | 1 | 0.43 | 0.13 | 5.13 | 7.36 | 2.18 | 2.79 | 1.67 | {100, 120, 140, 160} |
| | 2 | 2 | 0.49 | 0.14 | 6.88 | 9.88 | 2.04 | 1.94 | 1.46 | |
| | 3 | 3 | 0.09 | 0.03 | 5.04 | 7.24 | 2.11 | 1.64 | 1.00 | |
| | 4 | 4 | 0.47 | 0.14 | 6.37 | 9.14 | 2.62 | 1.88 | 1.13 | |
| | 5 | 5 | 0.13 | 0.04 | 6.57 | 9.43 | 2.94 | 1.41 | 1.10 | |
| | 6 | 6 | 0.20 | 0.06 | 6.07 | 8.71 | 2.35 | 1.55 | 0.98 | |
| | 7 | 7 | 0.05 | 0.01 | 6.77 | 9.72 | 2.41 | 1.15 | 0.88 | |
| | 8 | 8 | 0.04 | 0.01 | 6.80 | 9.76 | 2.98 | 0.91 | 0.78 | |
| | 9 | 9 | 0.11 | 0.03 | 6.25 | 8.97 | 2.95 | 0.92 | 1.10 | |
| | 10 | 10 | 0.07 | 0.02 | 5.28 | 7.58 | 2.68 | 0.85 | 0.97 | |
| CS5 | 1 | 3 | 0.86 | 0.26 | 5.13 | 7.36 | 2.18 | 2.79 | 1.67 | {50, 60, 70, 80} |
| | 2 | 3 | 0.98 | 0.28 | 6.88 | 9.88 | 2.04 | 1.94 | 1.46 | |
| | 3 | 2 | 0.65 | 0.06 | 5.04 | 7.24 | 2.11 | 1.64 | 1.00 | |
| | 4 | 2 | 0.57 | 0.28 | 6.37 | 9.14 | 2.62 | 1.88 | 1.13 | |
| | 5 | 4 | 0.63 | 0.18 | 6.57 | 9.43 | 2.94 | 1.41 | 1.10 | |
| | 6 | 4 | 0.70 | 0.22 | 6.07 | 8.71 | 2.35 | 1.55 | 0.98 | |
| | 7 | 2 | 0.55 | 0.12 | 6.77 | 9.72 | 2.41 | 1.15 | 0.88 | |
| | 8 | 2 | 0.54 | 0.12 | 6.80 | 9.76 | 2.98 | 0.91 | 0.78 | |
| | 9 | 3 | 0.61 | 0.16 | 6.25 | 8.97 | 2.95 | 0.92 | 1.10 | |
| | 10 | 3 | 0.57 | 0.14 | 5.28 | 7.58 | 2.68 | 0.85 | 0.97 | |
| CS6 | 1 | 2 | 0.67 | 0.07 | 6.61 | 19.09 | 2.96 | 1.97 | 1.07 | {80, 100, 120, 140} |
| | 2 | 2 | 0.61 | 0.06 | 6.88 | 19.87 | 2.27 | 1.09 | 0.32 | |
| | 3 | 3 | 0.32 | 0.03 | 5.68 | 16.41 | 2.63 | 1.73 | 1.26 | |
| | 4 | 3 | 0.63 | 0.06 | 6.94 | 20.05 | 2.87 | 1.06 | 1.72 | |
| | 5 | 2 | 0.59 | 0.06 | 5.15 | 14.88 | 2.68 | 1.45 | 0.82 | |
| | 6 | 2 | 0.45 | 0.05 | 6.28 | 18.14 | 2.49 | 1.61 | 0.63 | |
| | 7 | 1 | 0.42 | 0.04 | 5.64 | 16.29 | 2.85 | 2.05 | 0.41 | |
| | 8 | 1 | 0.60 | 0.06 | 5.06 | 14.62 | 2.79 | 1.87 | 1.96 | |
| | 9 | 2 | 0.48 | 0.05 | 5.61 | 16.21 | 2.31 | 2.41 | 1.47 | |
| | 10 | 2 | 0.52 | 0.05 | 5.84 | 16.87 | 2.92 | 0.87 | 0.83 | |
| | 11 | 3 | 0.35 | 0.04 | 6.71 | 19.38 | 2.94 | 0.92 | 1.11 | |
| | 12 | 3 | 0.30 | 0.03 | 5.58 | 16.12 | 2.64 | 1.80 | 0.74 | |
| | 13 | 2 | 0.67 | 0.07 | 5.68 | 16.41 | 2.60 | 2.47 | 0.60 | |
| | 14 | 2 | 0.57 | 0.06 | 5.19 | 14.99 | 2.30 | 1.63 | 1.81 | |
| | 15 | 1 | 0.37 | 0.04 | 6.09 | 17.59 | 2.72 | 1.39 | 0.93 | |

### *4.2. Solution encoding and initialization*

Two encoding schemes were employed for representing solutions to the Redundancy Allocation Problem (RAP): binary encoding and real number encoding, whose conceptual diagrams are shown in Fig. 7.

Binary encoding is designed for evolutionary algorithms, where candidate solutions are represented as chromosomes composed of binary genes (0 or 1). Each solution consists of a number of chromosomes corresponding to the number of subsystems in the overall system. Each chromosome includes 8 genes that encode the number of additional components in a given subsystem, along with 2 genes that represent the redundancy strategy. Four redundancy strategies were considered in this study, encoded as follows: 00 – cold standby, 01 – warm standby, 10 – mixed standby, and 11 – hot standby. This encoding scheme ensures that similar strategies occupy adjacent regions in the search space, thereby allowing smooth transitions between them with a monotonic increase in redundancy activation levels.

Real number encoding is intended for swarm intelligence-based algorithms and related metaheuristics, where the position of a particle in a multidimensional search space is defined by vectors of real numbers. One vector contains values in the range [0, 255] which encode the number of redundant components, while another vector contains values in the range [−0.5, 3.5] which encode the redundancy strategy. The redundancy strategy is determined by rounding the particle's coordinate to the nearest integer. The order of strategy representation in the search space is consistent with that used in the binary encoding, enabling coherent behavior across different algorithmic approaches and facilitating convergence through a structured and interpretable solution space.

For the initialization of the initial population in binary encoding, two methods were applied: Random Initialization (RI) and Scaled Binomial Initialization (SBI). Although RI is widely used across various metaheuristic algorithms due to its simplicity and implementation ease, SBI demonstrates notable advantages in large-scale optimization problems by introducing structured diversification into the initial population. In SBI, binary genes are generated with a variable probability of assigning 0 or 1, depending on the index of the individual solution within the initial population. This mechanism promotes both exploration and guided distribution of solutions in the early stages of the search process [55]. In contrast, for real number encoding, only Random Initialization is applicable due to the continuous nature of the encoding scheme and the absence of a straightforward probabilistic structure analogous to SBI for binary vectors.

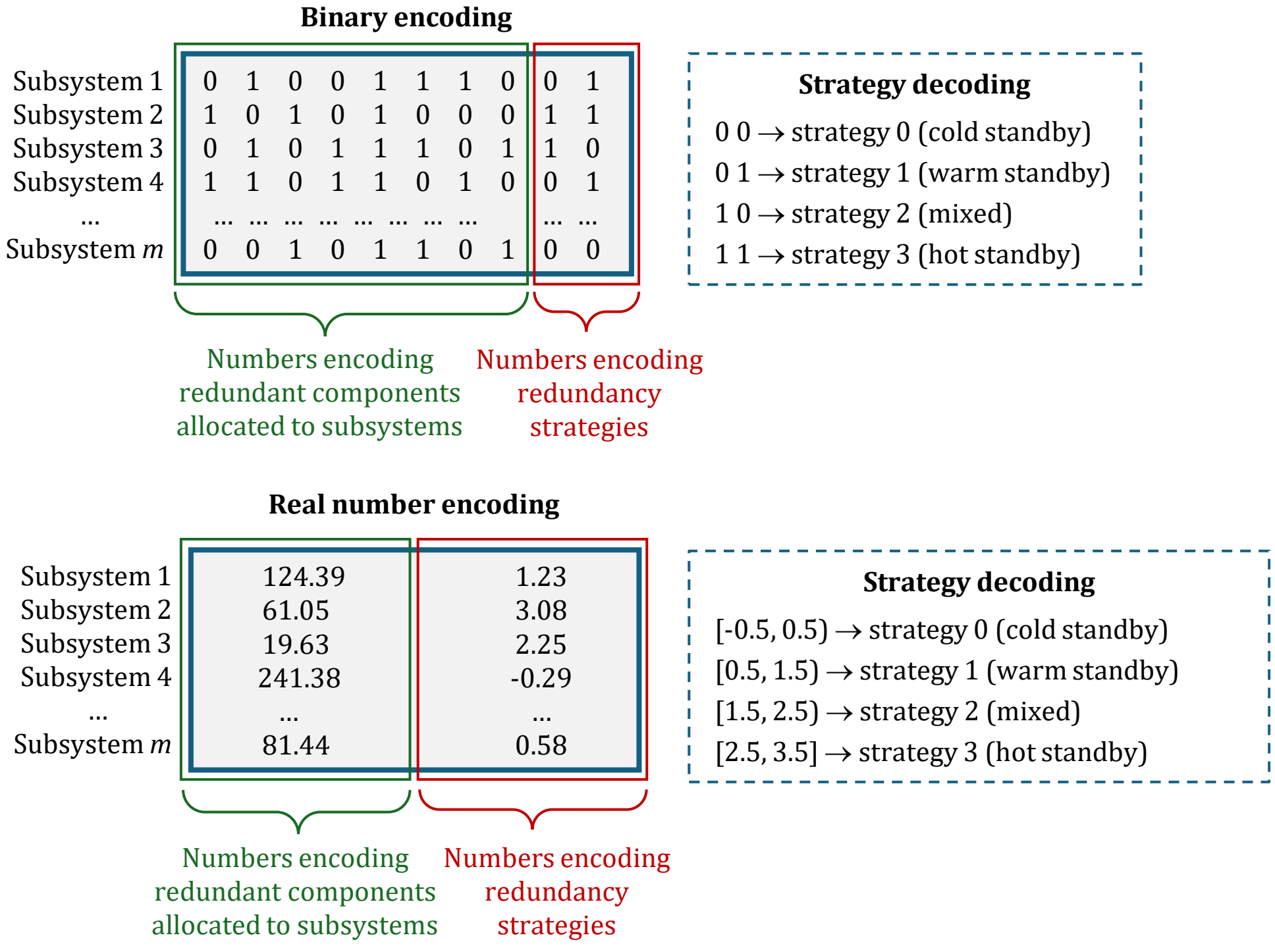

**Fig. 7.** Binary and real number encoding for standby components and redundancy strategies.

### *4.3. Metaheuristic algorithms used in the benchmark study*

For the selection of the metaheuristic algorithms, we utilized the PlatEMO platform [64], one of the largest and most widely used open-source platforms for evolutionary multi-objective optimization. PlatEMO contains more than 300 implementations of standard and state-of-the-art algorithms, provides default values for algorithm-specific parameters, and supports efficient large-scale benchmarking studies [65]. From this pool, we selected 65 algorithms that were compatible with the implemented RAP formulation and successfully passed preliminary execution tests. These algorithms are summarized in Table 3 and classified into ten groups, as illustrated in Fig. 8.

The algorithm selection was designed to provide broad coverage of the multi-objective soft computing optimizer landscape rather than to promote a single algorithmic paradigm. The inclusion criteria were as follows: the algorithm had to be applicable to multi-objective optimization, compatible with the implemented mixed-integer RAP encoding after discretization of the decision variables, executable within the adopted PlatEMO workflow, and capable of producing valid candidate solutions in preliminary test runs. Algorithms were excluded if they required unsupported problem-specific operators, assumptions incompatible with the discrete constrained RAP formulation, unavailable derivative information, nonstandard external data, were not designed to run on large computational budgets (by utilizing surrogates) or if they repeatedly failed to produce valid outputs in pilot experiments.

The resulting algorithm portfolio includes representatives of several major families, including evolutionary, swarm-based, decomposition-based, indicator-based, dominance-based, differential-evolution-based, reference-vector-based, coevolutionary, multitasking, and adaptive methods. All algorithms were compared under the same population size, termination criterion, objective-function evaluation budget, constraint-handling procedure, initialization settings, and performance indicators. Default PlatEMO parameter settings were retained unless minor compatibility adjustments were required. This choice reflects a practical benchmarking perspective: the aim is to assess standardized out-of-the-box robustness and budget-dependent behavior rather than individually tuned performance of each algorithm. Additionally, as the performed computational experiments were extensive (as reported in the following section), the addition of hyperparameter optimization for all the algorithms would be prohibitively computationally expensive.

Consequently, the results should not be interpreted as proving the intrinsic superiority of any algorithm under optimal parameter tuning. Instead, they indicate which methods and algorithmic families perform reliably under the adopted RAP encoding, initialization schemes, constraints, and computational budgets.

The classification in Fig. 8 is used to support the interpretation of the benchmark results at the algorithm-family level. Evolutionary and differential-evolution-based algorithms mainly rely on population variation operators such as selection, crossover, mutation, and vector-difference-based perturbation. Swarm-based algorithms exploit collective search mechanisms inspired by social or biological behavior. Decomposition-based and reference-vector-based algorithms structure the search through scalar subproblems, reference directions, or regions of the objective space. Indicator-based and dominance-based algorithms guide selection using quality indicators or Pareto-dominance relations. Coevolutionary, multitasking, multi-stage, and adaptive algorithms introduce additional mechanisms for population interaction, knowledge transfer, staged search, or dynamic adjustment of the search process. This grouping enables the subsequent analysis to move beyond individual algorithm rankings and identify broader performance patterns across algorithmic families.

**Table 3**
List of algorithms used in the study.

| Abbreviation | Name | Ref. | Abbreviation | Name | Ref. |
|---|---|---|---|---|---|
| ANSGA-III | Adaptive Non-dominated Sorting Genetic Algorithm III | [66] | MSCEA | Multi-stage Constrained Multi-objective Evolutionary Algorithm | [67] |
| AGE-MOEA | Adaptive Geometry Estimation-based Many-objective Evolutionary Algorithm | [68] | MSCMO | Multi-stage Constrained Multi-objective Evolutionary Algorithm | [69] |
| AGE-MOEA-II | Adaptive Geometry Estimation-based Many-objective Evolutionary Algorithm II | [70] | MTCMO | Multitasking Constrained Multi-objective Optimization | [71] |
| AR-MOEA | Adaptive Reference-guided Many-objective Evolutionary Algorithm | [72] | MaOEA-DDFC | Many-objective Evolutionary Algorithm based on Directional Diversity and Favorable Convergence | [73] |
| BCE-IBEA | Bi-Criterion Evolution Indicator-based Evolutionary Algorithm | [74] | MyO-DEMR | Many-objective Differential Evolution with Mutation Restriction | [75] |

| Abbreviation | Name | Ref. | Abbreviation | Name | Ref. |
|---|---|---|---|---|---|
| BCE-MOEA/D | Bi-Criterion Evolution Indicator-based MOEA/D | [74] | NMPSO | Novel MOPSO | [76] |
| BiCo | Bidirectional Coevolution Constrained Multi-objective Evolutionary Algorithm | [77] | NNIA | Non-dominated Neighbor Immune Algorithm | [78] |
| C3M | Constraint, Multi-objective, Multi-stage, Multi-constraint Evolutionary Algorithm | [79] | NRV-MOEA | Adaptive Normal Reference Vector-based Multi- and Many-objective Evolutionary Algorithm | [80] |
| CA-MOEA | Clustering-based Adaptive Multi-objective Evolutionary Algorithm | [81] | NSBiDiCo | Non-dominated Sorting Bidirectional Differential Coevolution Algorithm | [82] |
| CAEAD | Dual-population Evolutionary Algorithm Based on Alternative Evolution and Degeneration | [83] | NSGA-II | Non-dominated Sorting Genetic Algorithm II | [84] |
| CCMO | Coevolutionary Constrained Multi-objective Optimization Framework | [85] | NSGA-II+ARSBX | NSGA-II with Adaptive Rotation based Simulated Binary Crossover | [86] |
| CLIA | Evolutionary Algorithm with Cascade Clustering and Reference Point Incremental Learning | [87] | NSGA-III | Non-dominated Sorting Genetic Algorithm III | [88] |
| CMEGL | Constrained Evolutionary Multitasking with Global and Local Auxiliary Tasks | [89] | NSLS | Multi-objective Optimization Framework based on Non-dominated Sorting and Local Search | [90] |
| CMODE-FTR | Constrained Multi-objective Differential Evolution Based on the Fusion of Two Rankings | [91] | PESA-II | Pareto Envelope-based Selection Algorithm II | [92] |
| CMOEA-MS | Constrained Multi-objective Evolutionary Algorithm with Multiple Stages | [93] | PICEA-g | Preference-inspired Coevolutionary Algorithm with Goals | [94] |
| CMOEA-MSG | Multi-stage Constrained Multi-objective Evolutionary Algorithm | [95] | RM-MEDA | Regularity Model-based Multi-objective Estimation of Distribution | [96] |
| CMOPSO | Competitive Mechanism based MOPSO | [97] | RSEA | Radial Space Division based Evolutionary Algorithm | [98] |
| DCNSGA-III | Dynamic Constrained Non-dominated Sorting Genetic Algorithm III | [99] | RVEA | Reference Vector Guided Evolutionary Algorithm | [100] |
| DEA-GNG | Decomposition Based Evolutionary Algorithm Guided by Growing Neural Gas | [101] | RVEA-iGNG | RVEA based on Improved Growing Neural Gas | [102] |
| DRLOS-EMCMO | EMCMO with Deep Reinforcement Learning-assisted Operator Selection | [103] | RVEAa | RVEA Embedded with the Reference Vector Regeneration Strategy | [100] |
| DSPCMDE | Dynamic Selection Preference-assisted Constrained Multi-objective Differential Evolution | [104] | SPEA-R | Strength Pareto Evolutionary Algorithm based on Reference Direction | [105] |
| EFR-RR | Ensemble Fitness Ranking with a Ranking Restriction Scheme | [106] | SPEA2 | Strength Pareto Evolutionary Algorithm 2 | [107] |
| EMyO-C | Evolutionary Many-objective Optimization Algorithm with Clustering-based Selection | [108] | SSCEA | Subspace Segmentation based Co-evolutionary Algorithm | [109] |
| GDE3 | Generalized Differential Evolution 3 | [110] | TS-NSGA-II | Two-stage NSGA-II | [111] |
| IBEA | Indicator-based Evolutionary Algorithm | [112] | TSTI | Two-stage Evolutionary Algorithm with Three Indicators | [113] |
| ICMA | Indicator-based Constrained Multi-objective Algorithm | [114] | ToP | Two-phase Framework with NSGA-II | [115] |
| IMTCMO | Improved Evolutionary Multitasking-based CMOEA | [116] | dMOPSO | MOPSO based on Decomposition | [117] |
| MCCMO | Multi-population Coevolutionary Constrained Multi-objective Optimization | [118] | $g$-NSGA-II | $g$-Dominance based NSGA-II | [119] |
| MFO-SPEA2 | Multiform Optimization Framework based on SPEA2 | [120] | hpaEA | Hyperplane Assisted Evolutionary Algorithm | [121] |
| MMOPSO | MOPSO with multiple search strategies | [122] | one-by-one EA | Many-objective Evolutionary Algorithm using a one-by-one Selection Strategy | [123] |
| MOBCA | Multi-objective Besiege and Conquer Algorithm | [124] | $\theta$-DEA | Theta-Dominance based Evolutionary Algorithm | [125] |
| MOEA/D-CMA | MOEA/D with Covariance Matrix Adaptation Evolution Strategy | [126] | $\theta$-DEA-CPBI | Theta-Dominance based Evolutionary Algorithm with Constrained Penalty Boundary Intersection | [127] |
| MOPSO-CD | MOPSO with Crowding Distance | [128] | | | |

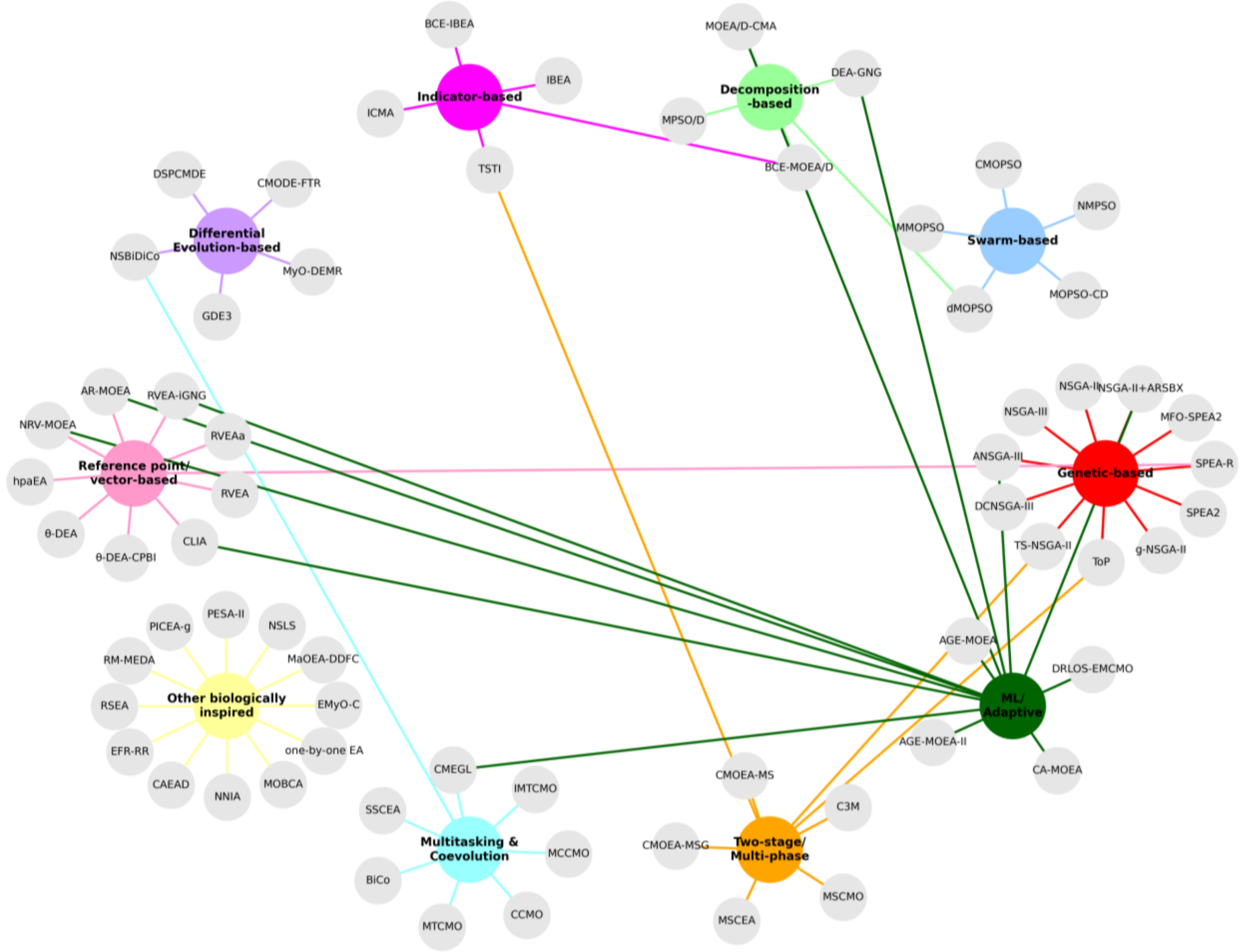

**Fig. 8.** Classification of algorithms used in the study.

### *4.4. Experimental setup*

The PlatEMO framework was used as the primary source for the implementation of the algorithms (described in Table 3). Each of the 65 algorithms was run in default setting defined by the PlatEMO framework, with one exception – the population size for all methods was set to 200. All the algorithms were then run 10 times on the all problems (24 in total: 6 case studies times 4 values of $W$) in two versions – one that used SBI and one that did not. The total budget of function evaluations was set to 2e6 (which corresponded to 1e4 iterations of algorithms with a fixed population size). In total, 31200 experiments were run. Depending on the algorithm, these runs took anywhere between tens of minutes to several hours. Because of the extensive computational burden, all the computational experiments were performed on the MetaCentrum cluster[1]. The codes used to run the experiments and the resulting aggregate data are available at a Zenodo repository[2]. The full data from the experiments (roughly 500GB) were too large to upload into the public repository.

## 5. Results and discussion

As the RAP formulation is bi-objective (minimizing cost and maximizing availability), we will utilize an indicator called Hypervolume (HV) to compare the solutions obtained by the individual algorithms [129]. This measure uses a reference point to calculate the area of the best-found non-dominated solutions, as depicted in Fig. 9.

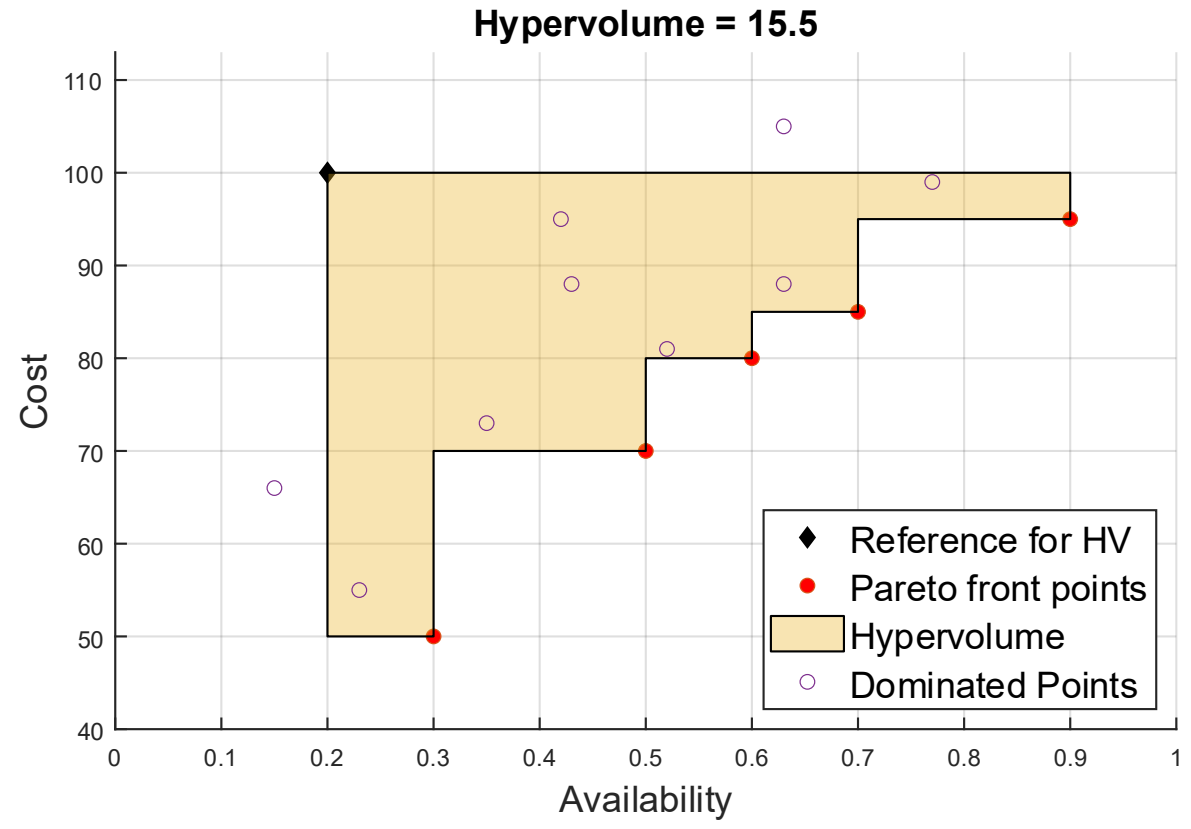

**Fig. 9.** Depiction of the Hypervolume indicator on a hypothetical example.

[1] https://metavo.metacentrum.cz/en/index.html
[2] 10.5281/zenodo.17981721

### 5.1. Redundancy-strategy composition of final populations and Pareto fronts

First, we investigate the **RQ1**, i.e. finding out type of redundancy strategy is most commonly preferred by subsystems. To do this, we look at the final populations of all considered methods. Here, we look at the different proportions of the redundancy strategies in the different subsystems. Additionally, we look at only the Pareto-optimal solutions that are aggregated over all final populations. This was done by putting together all final populations for all methods over all runs for a given CS/$W$ combination. From this large pool of solutions, the Pareto-optimal ones were filtered.

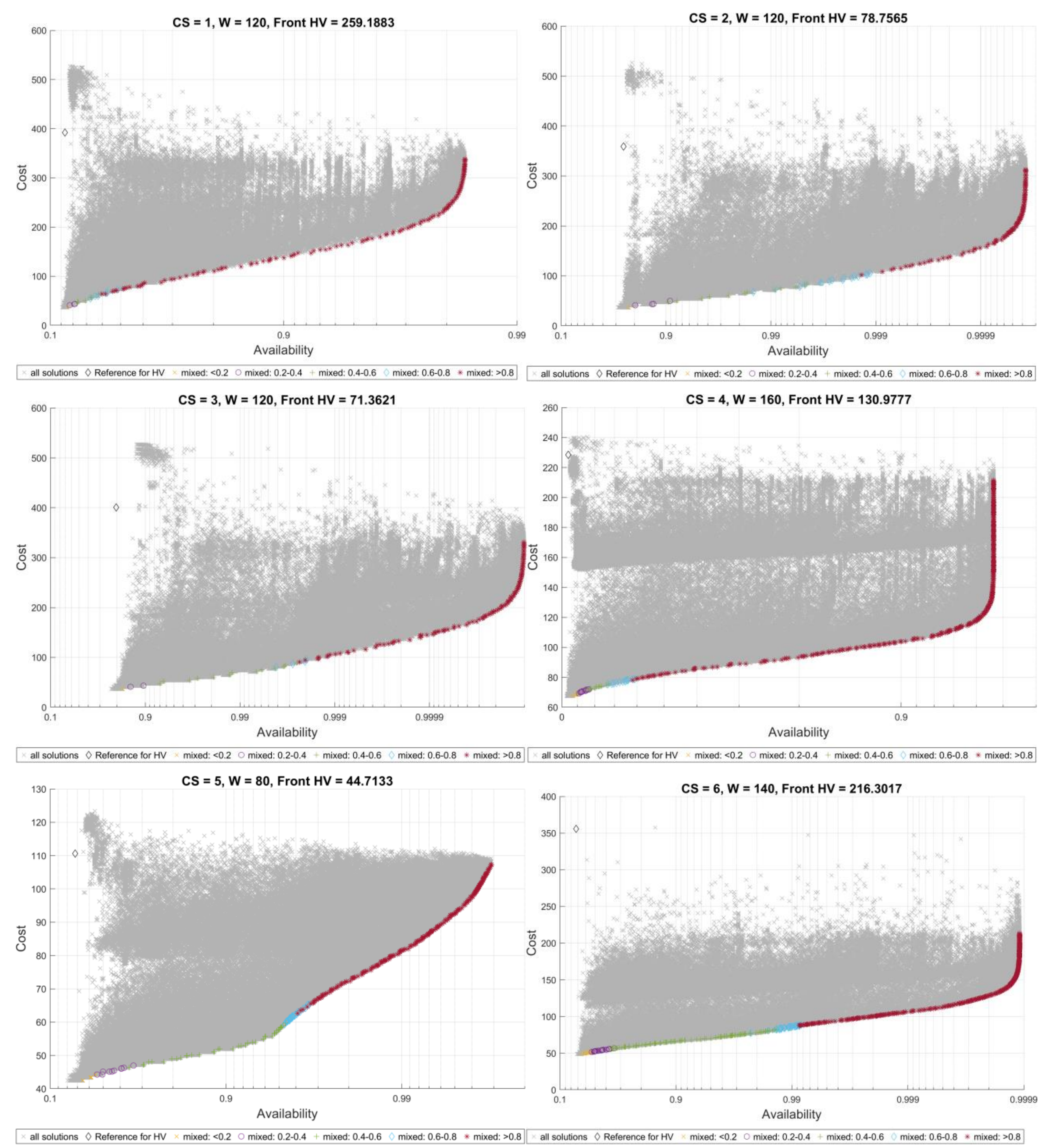

**Fig. 10.** Solutions from all populations, different CS with maximum value of $W$. Pareto-optimal points are highlighted and categorized by the proportion of the mixed strategy.

**Table 4**

Proportion of different strategies in final populations of the algorithms (All solutions and Pareto-optimal ones).

| | | *W*=60 | | *W*=80 | | *W*=100 | | *W*=120 | |
|---|---|---|---|---|---|---|---|---|---|
| | **strategy** | All | Pareto | All | Pareto | All | Pareto | All | Pareto |
| **CS1** | cold | 0.01 | 0.00 | 0.01 | 0.00 | 0.01 | 0.00 | 0.01 | 0.00 |
| | warm | 0.02 | 0.00 | 0.02 | 0.00 | 0.02 | 0.00 | 0.02 | 0.00 |
| | mixed | 0.74 | 0.78 | 0.79 | 0.82 | 0.81 | 0.82 | 0.82 | 0.84 |
| | hot | 0.23 | 0.22 | 0.19 | 0.18 | 0.17 | 0.18 | 0.15 | 0.16 |

| | | *W*=60 | | *W*=80 | | *W*=100 | | *W*=120 | |
|---|---|---|---|---|---|---|---|---|---|
| | **strategy** | All | Pareto | All | Pareto | All | Pareto | All | Pareto |
| **CS2** | cold | 0.02 | 0.00 | 0.02 | 0.00 | 0.01 | 0.00 | 0.01 | 0.00 |
| | warm | 0.02 | 0.00 | 0.02 | 0.00 | 0.02 | 0.00 | 0.02 | 0.00 |
| | mixed | 0.48 | 0.48 | 0.54 | 0.51 | 0.59 | 0.54 | 0.61 | 0.56 |
| | hot | 0.48 | 0.52 | 0.43 | 0.49 | 0.38 | 0.46 | 0.36 | 0.44 |

| | | *W*=60 | | *W*=80 | | *W*=100 | | *W*=120 | |
|---|---|---|---|---|---|---|---|---|---|
| | **strategy** | All | Pareto | All | Pareto | All | Pareto | All | Pareto |
| **CS3** | cold | 0.02 | 0.00 | 0.01 | 0.00 | 0.01 | 0.00 | 0.01 | 0.00 |
| | warm | 0.02 | 0.00 | 0.02 | 0.00 | 0.02 | 0.00 | 0.02 | 0.00 |
| | mixed | 0.54 | 0.50 | 0.60 | 0.55 | 0.63 | 0.57 | 0.65 | 0.58 |

| | | *W*=100 | | *W*=120 | | *W*=140 | | *W*=160 | |
|---|---|---|---|---|---|---|---|---|---|
| | **strategy** | All | Pareto | All | Pareto | All | Pareto | All | Pareto |
| **CS4** | cold | 0.01 | 0.00 | 0.01 | 0.00 | 0.00 | 0.00 | 0.01 | 0.00 |
| | warm | 0.01 | 0.00 | 0.01 | 0.00 | 0.01 | 0.00 | 0.01 | 0.00 |
| | mixed | 0.80 | 0.82 | 0.83 | 0.86 | 0.85 | 0.83 | 0.84 | 0.85 |

| | strategy | All | Pareto | All | Pareto | All | Pareto | All | Pareto |
|---|---|---|---|---|---|---|---|---|---|
| | hot | 0.43 | 0.50 | 0.37 | 0.45 | 0.34 | 0.43 | 0.32 | 0.42 |
| | | *W*=50 | | *W*=60 | | *W*=70 | | *W*=80 | |
| | **strategy** | All | Pareto | All | Pareto | All | Pareto | All | Pareto |
| **CS5** | cold | 0.01 | 0.00 | 0.01 | 0.00 | 0.01 | 0.00 | 0.01 | 0.00 |
| | warm | 0.01 | 0.00 | 0.01 | 0.00 | 0.01 | 0.00 | 0.01 | 0.00 |
| | mixed | 0.42 | 0.40 | 0.60 | 0.54 | 0.64 | 0.54 | 0.67 | 0.55 |
| | hot | 0.56 | 0.60 | 0.38 | 0.46 | 0.34 | 0.46 | 0.32 | 0.45 |

| | strategy | All | Pareto | All | Pareto | All | Pareto | All | Pareto |
|---|---|---|---|---|---|---|---|---|---|
| | hot | 0.18 | 0.18 | 0.15 | 0.14 | 0.14 | 0.17 | 0.14 | 0.15 |
| | | *W*=80 | | *W*=100 | | *W*=120 | | *W*=140 | |
| | **strategy** | All | Pareto | All | Pareto | All | Pareto | All | Pareto |
| **CS6** | cold | 0.01 | 0.00 | 0.01 | 0.00 | 0.00 | 0.00 | 0.00 | 0.00 |
| | warm | 0.02 | 0.00 | 0.02 | 0.00 | 0.01 | 0.00 | 0.01 | 0.00 |
| | mixed | 0.56 | 0.50 | 0.61 | 0.51 | 0.63 | 0.50 | 0.63 | 0.48 |
| | hot | 0.42 | 0.50 | 0.37 | 0.49 | 0.36 | 0.50 | 0.36 | 0.52 |

Fig. 10 presents all final-population solutions in the objective space together with the computed aggregated Pareto fronts, while Table 4 reports the proportions of redundancy strategies among all final solutions and Pareto-optimal solutions. Cold standby and warm standby occur only marginally in the final populations and are absent from the aggregated Pareto fronts. The Pareto-optimal designs are therefore composed almost exclusively of hot standby and active–warm mixed configurations. Across most case studies and weight limits, the active–warm mixed strategy has the largest share, and its proportion generally increases as the weight constraint is relaxed. From a system-design perspective, this pattern is consistent with the adopted CTMC-based availability model and parameterization. Under tight weight constraints, only a limited number of spare components can be allocated across subsystems. In this situation, hot standby is attractive because it provides immediate functional substitution after the failure of an active component, giving it an advantage over cold and warm standby configurations. As the weight limit increases, more spare components can be deployed. When a subsystem contains several spares, keeping all of them in hot standby increases their exposure to active-mode failure intensity while providing limited additional benefit beyond immediate switchover. In such cases, the active–warm mixed strategy becomes more advantageous: one spare component is maintained in active mode, while the remaining spares are kept in warm standby. This configuration provides a compromise between reducing downtime caused by an insufficient number of operating components and limiting the degradation and failure intensity of spare components.

The absence of cold and warm standby from the aggregated Pareto fronts should also be interpreted in relation to the adopted model assumptions. Cold standby avoids standby-state failures before activation, but its benefit is offset by the slower activation process. Warm standby enables faster activation, but standby components remain exposed to failure. Under the current CTMC assumptions and parameter values, these mechanisms make cold and warm standby less competitive in the cost–availability objective space than hot standby and the active–warm mixed strategy. These conclusions should not be generalized to all repairable systems or all standby policies. They are conditional on the adopted CTMC formulation, the transition-rate and cost/weight parameterization, and the specific active–warm mixed strategy considered in this benchmark. Different switching rates, standby failure rates, repair rates, imperfect switching, common-cause failures, non-exponential transition times, or alternative mixed redundancy policies could change the relative attractiveness of the strategies. As also shown in Fig. 10, CS1 and CS4 do not achieve high-availability solutions above 0.99. The shapes of the aggregated Pareto fronts differ across case studies. In CS2, CS3, and CS6, the fronts are convex. In CS1 and CS4, the fronts initially have a concave shape and become convex for higher-cost and higher-availability solutions. In CS5, the front is more irregular and alternates between convex and concave regions.

### *5.2. Hypervolume-based algorithm performance*

To answer **RQ2-3**, we implemented several approaches for the comparison of the algorithms. For each problem (a combination of CS and *W*, 24 different ones) we computed the HV of the Pareto-optimal solutions found by the individual algorithms for different computation budgets (between 2e2 and 2e6) in each run. As the reference point for the HV, we selected the worst values of any Pareto-optimal solution found by any algorithm with the full 2e6 budget (and this reference point was then the same for all HV computation for that problem). Fig. 10 shows the positions of these reference points for several problems. Naturally, the choice of the point for the HV computation has direct impact on the assessment of the performance of the methods. We purposefully selected the points in a way that guarantees that whole Pareto front (approximated by the union of all considered algorithms) is covered in the computation. Moreover, as is visible from the examples in Fig. 10, this choice of points for the HV computation is robust to small changes, as is not directly on minimum availability/maximum cost values of the solutions from the Pareto front.

We can then look at the convergence of HV for the individual methods on the different problems, which is demonstrated on Fig. 11 (with several methods, that appeared as top-performing ones based on various criteria discussed further, being highlighted). Here, we can make several observations. The first is the superiority of the SBI initialization - methods that utilized SBI get a significant head-start. For CS2 and CS3, the SBI initialization alone produced solution sets that were at just 10% off from the best found HV. Many methods converge to the same HV value, but the speed at which they converge varies quite widely. For the smaller-dimension CS1-CS3, roughly 1e4 evaluations were needed for many methods to converge close to the best-found HV. For the larger-dimension CS4-CS6, it was roughly 1e5 evaluations. For CS1-CS3, we can see that many variants of the algorithms without SBI could eventually converge close to the best-found HV. However, for CS4-CS6, this was no longer the case – here, only a handful of methods without SBI could find very good solution sets.

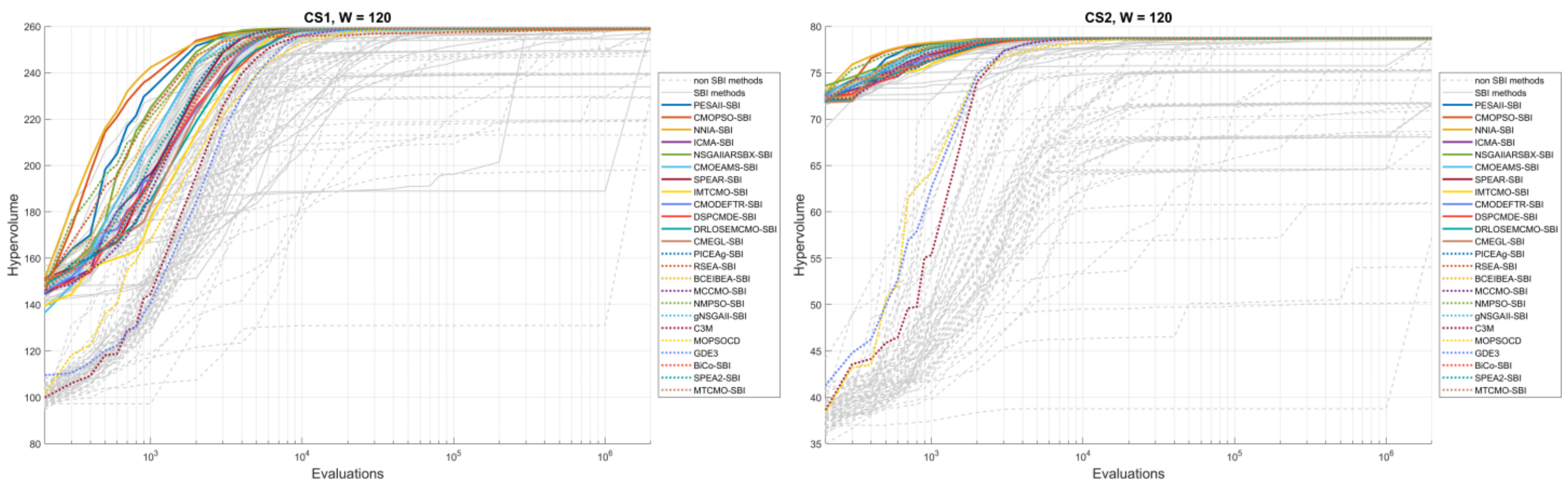

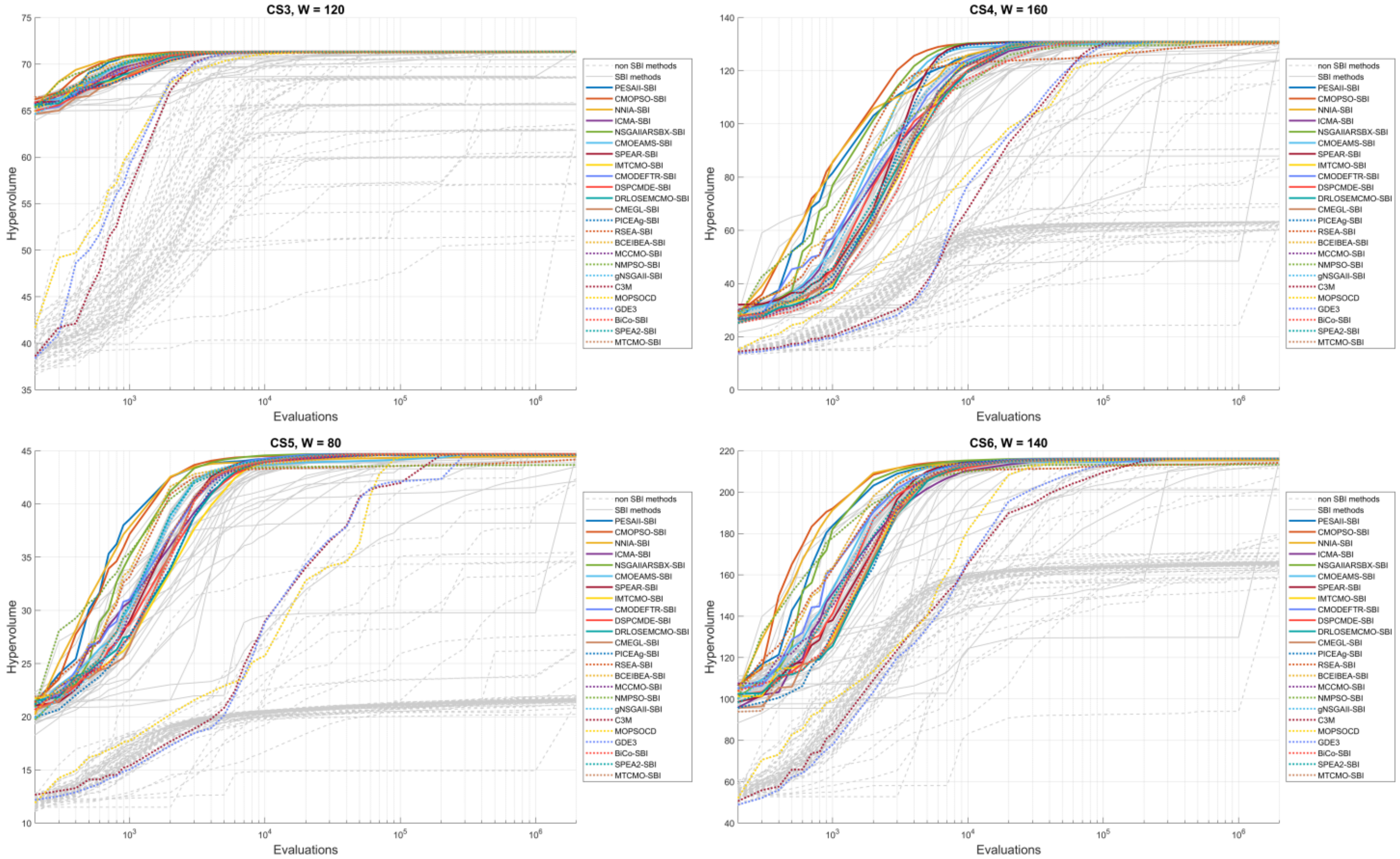

**Fig. 11.** Convergence of HV different methods on selected problems (average over the 10 runs). Selected methods are highlighted.

### *5.3. Budget-dependent statistical ranking of algorithms*

To make proper algorithmic comparison and find the best-performing methods for different computational budgets, we performed an in-depth statistical analysis of the results [130]. We followed the guidelines published in LaTorre et al. (2021) and Bartz-Beielstein et al. (2020). The procedure worked as follows:

- Based on the HV values of the different problems/budgets/runs, the Friedman ranks of the algorithms were computed.
- Then we utilized the Wilcoxon signed-rank test to find out whether statistically significant differences exist between the best algorithm (the one with the lowest Friedman rank in the given CS/$W$ combination) and the other methods.
- Once the pairwise p-values from the Wilcoxon test were computed, we applied the Holm-Bonferroni [132] correction method, which counteracts the effect of multiple comparisons by controlling the family-wise error rate [133].
- Finally, if the corrected p-value was below a threshold value of $\alpha = 0.05$ we denote the algorithm as being indistinguishable from the best algorithm on the given problem/budget setting.

The results of this comparison are summarized in heatmaps in Figs. 12 and 13, where we divided the algorithms into groups following the classification in Fig. 7. From these results, we can see that the classification is not a perfect predictor of success – in each class, there were methods that performed very poorly regardless of the computational budget. However, there are classes, such as DE-based, Multitask, ML-adaptive, and Two-stage, in which the majority of the methods performed quite well on various computational budgets.

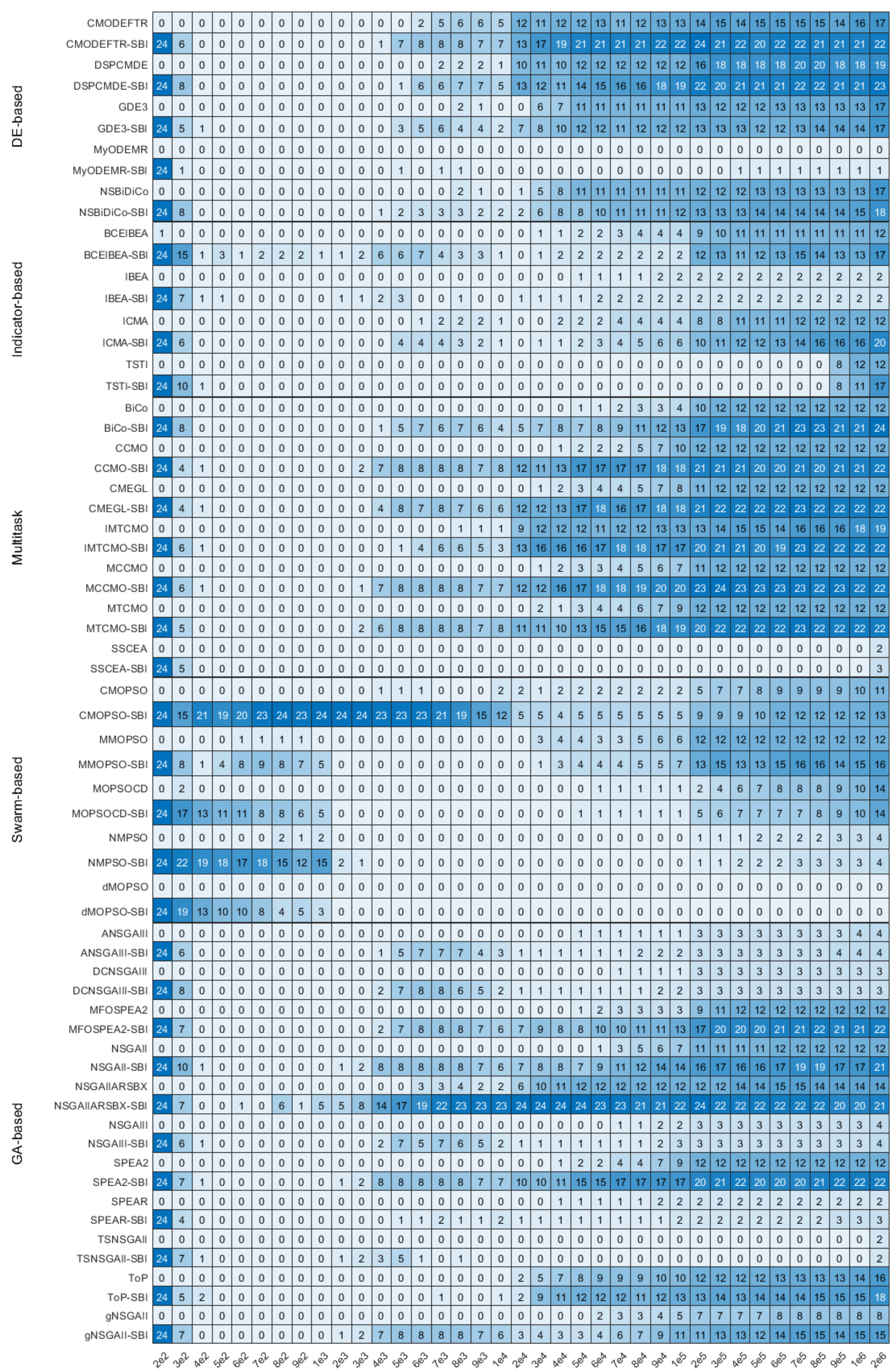

**Fig. 12.** Number of times the algorithm was indistinguishable from the best algorithm for a given problem (out of 24 problems in total). First half of considered methods.

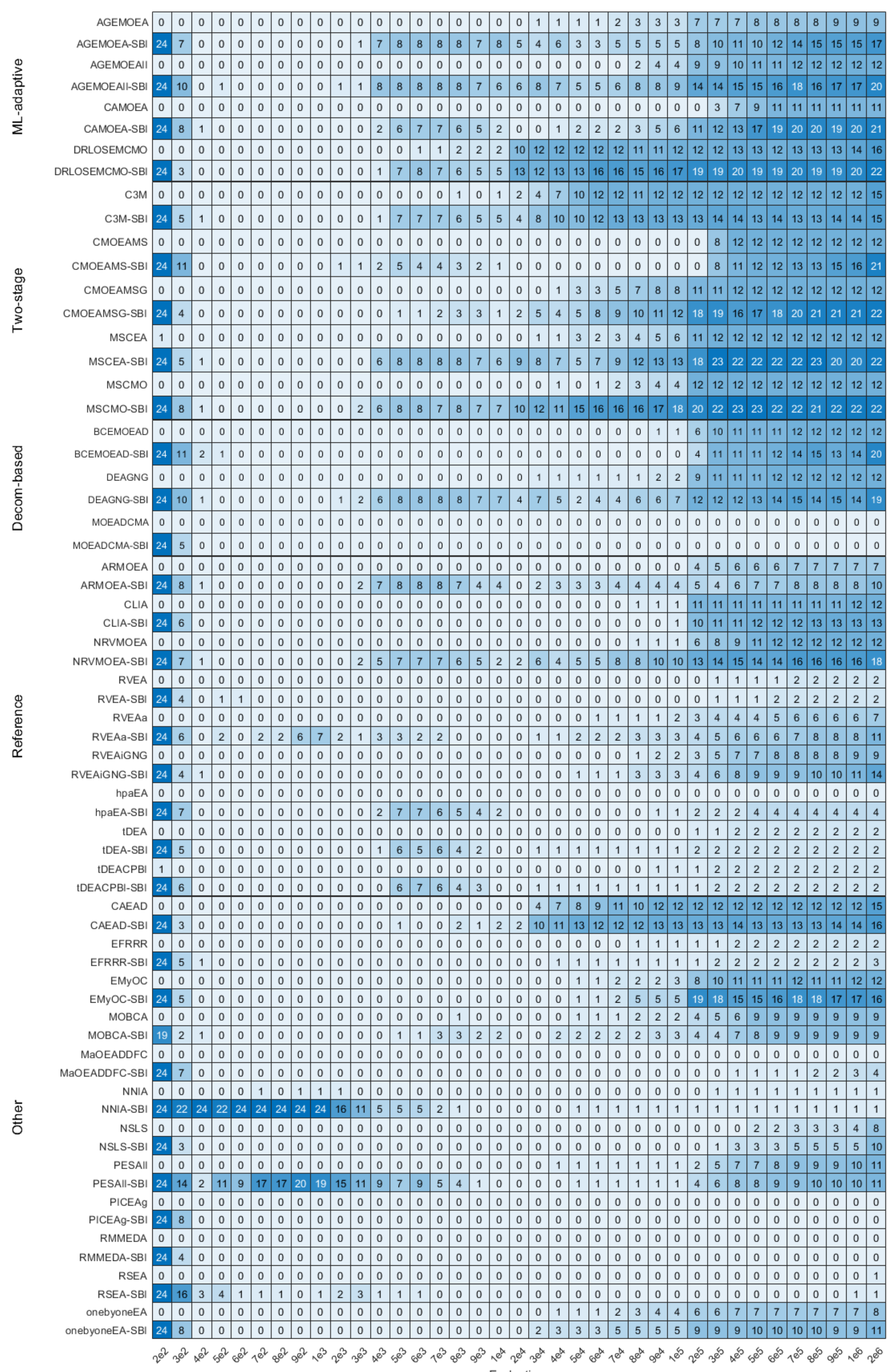

**Fig. 13.** Number of times the algorithm was indistinguishable from the best algorithm for a given problem (out of 24 problems in total). Second half of considered methods.

We will look at each class separately and point out some outstanding methods:

- **Differential Evolution-based**: CMODEFTR-SBI and DSPCMDE-SBI achieved great performance (indistinguishable from the best method on 75%, or 18/24, of the problems) for budgets from 2e4 and 9e4, respectively. Interestingly, when compared to other classes, for many methods in this class the performance of SBI and non-SBI variants for large computational budgets are not very different, especially for GDE3 and NSBiDiCo.
- **Indicator-based**: In this class, there are not many outstanding methods. The three that performed relatively well, BCEIBEA-SBI, ICMA-SBI, and TSTI-SBI, all "come online" only for very large computational budgets.
- **Multitask & Coevolution**: Here, almost all SBI variants of the methods, apart from SSCEA, could get great performance with computational budgets larger than 5e4.
- **Swarm-based**: The methods from this class dominated the low computational budget settings. Especially CMOPSO-SBI was one of the top methods for budgets between 4e2 and 7e3. However, its performance degraded quite quickly for the budgets between 1e4 and 1e5, but then recovered slightly in the largest budget settings.
- **Genetic-based**: This class was a "mixed-bag" of both some of the worst and best-performing methods. The outstanding one was NSGAIIARSBX-SBI, which kept great performance for the largest stretch of all considered methods – between 6e3 and 2e6 evaluations. Other notable methods were MFOSPEA2-SBI and the "traditional" methods SPEA2-SBI and NSGAII-SBI. Interestingly, NSGAIII, which is the "improved" version of NSGAII, performed very poorly in our experiments (both versions with and without SBI).
- **ML/adaptive**: DRLOSEMCMO-SBI and CAMOEA-SBI achieved great performance for large budgets from 1e5 and 6e5, respectively. Here, all methods (also the non-SBI ones) could be indistinguishable from the best method on a handful of problem/budget combinations.
- **Two-stage/Multi-phase**: Structurally very similar results as in the ML-adaptive class, with CMOEAMSG-SBI, MSCEA-SBI, and MSCMO-SBI achieving great performance for budgets larger than 2e5.
- **Decomposition-based**: Similar to the Indicator-based class, the two methods that performed relatively well, BCEMOEAD-SBI and DEAGNG-SBI, do so only on the largest computational budgets.
- **Reference point/vector-based**: Despite having relatively large number of methods, this is probably the class with the "least interesting" methods overall in terms of performance, without truly outstanding methods, apart from NRVMOEA-SBI for the largest budgets.
- **Other biologically inspired**: With the exception of EMyOC-SBI, methods in this class perform very poorly on large budgets. However, two methods, PESAII-SBI and especially NNIA-SBI, were among the very best ones for small budgets roughly until 2e3.

### *5.4. Relative-distance analysis to the best-found HV*

Although the statistical analysis on the ranking of the methods is valuable, it is still a bit limited in the information it provides. It tells if the method is among the best, but it does not tell us how good its solutions are with respect to the "optimal ones". In our case, as we do not have access to the "true" Pareto front, we will measure the relative distance between the HV of the solution set obtained by the given method (averaged over the 10 runs) on a specific computational budget to the HV of the best-found solution set of all considered methods on the full computational budget (2e6 evaluations) over all runs. We then average this number over all 24 problems. If a method has a relative distance to the best-found HV on a specific budget equal to $10^{-2}$ it means that on average (over the 24 problems and 10 runs) it produces solution sets that have 99.9% of the HV of the best-found solutions over all methods.

In Fig. 14 we can see the convergence of this relative distance for some selected methods. We can see that many methods can get to a relative distance of $10^{-2}$ within a 1e4 or 2e4 budget (and methods like CMOPSO-SBI or NSGAIIARSBX-SBI even sooner). The $10^{-3}$ relative distance is first crossed by NSGAIIARSBX-SBI at roughly the 2e4 budget and can be eventually achieved by almost all highlighted methods. The $10^{-4}$ relative distance can be achieved only by a handful of methods. In Fig. 14, we also show the convergence for two "hypothetical" methods – an average one and a virtual best one (i.e., picking for each problem and budget the method that has the lowest relative distance) – and divide them by the use of SBI. We can see that for the "average" method without SBI does not converge even over the $10^{-1}$ relative distance value. With SBI, the "average" method can converge just under the $10^{-2}$ relative distance. The "virtual best without SBI" starts off much slower (thanks to the SBI providing a substantial head-start), but can converge to relative distance values under $10^{-4}$. The "virtual best with SBI" is, naturally, above all other convergence curves and get to values under $10^{-5}$. Interestingly, there are budget ranges, where two methods, CMOPSO and NSGAIIARSBX-SBI, touch the "virtual best with SBI" line, signaling they are the best methods on all 24 problems on the specific budget.

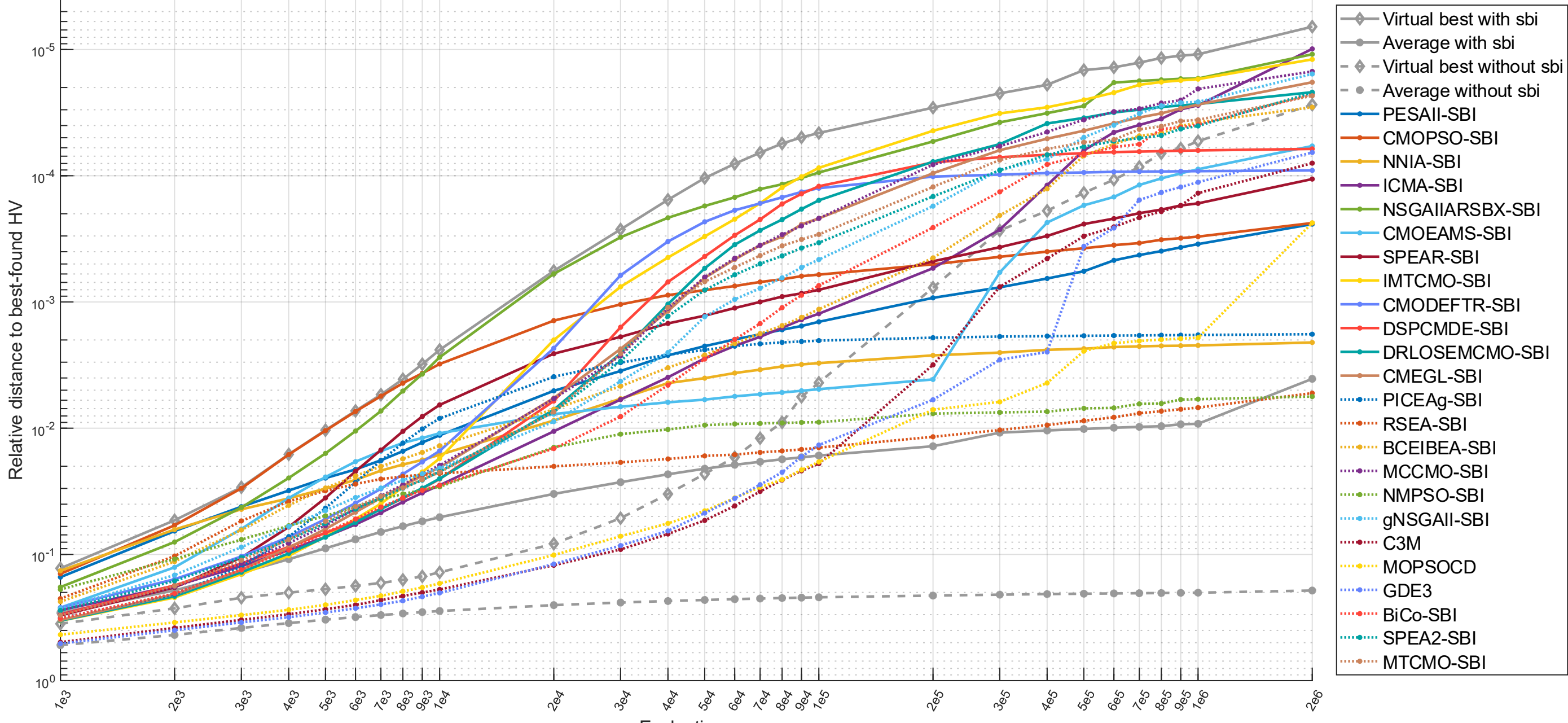

**Fig. 14.** Convergence of the relative distance from the best-found HV values over the 24 problems.

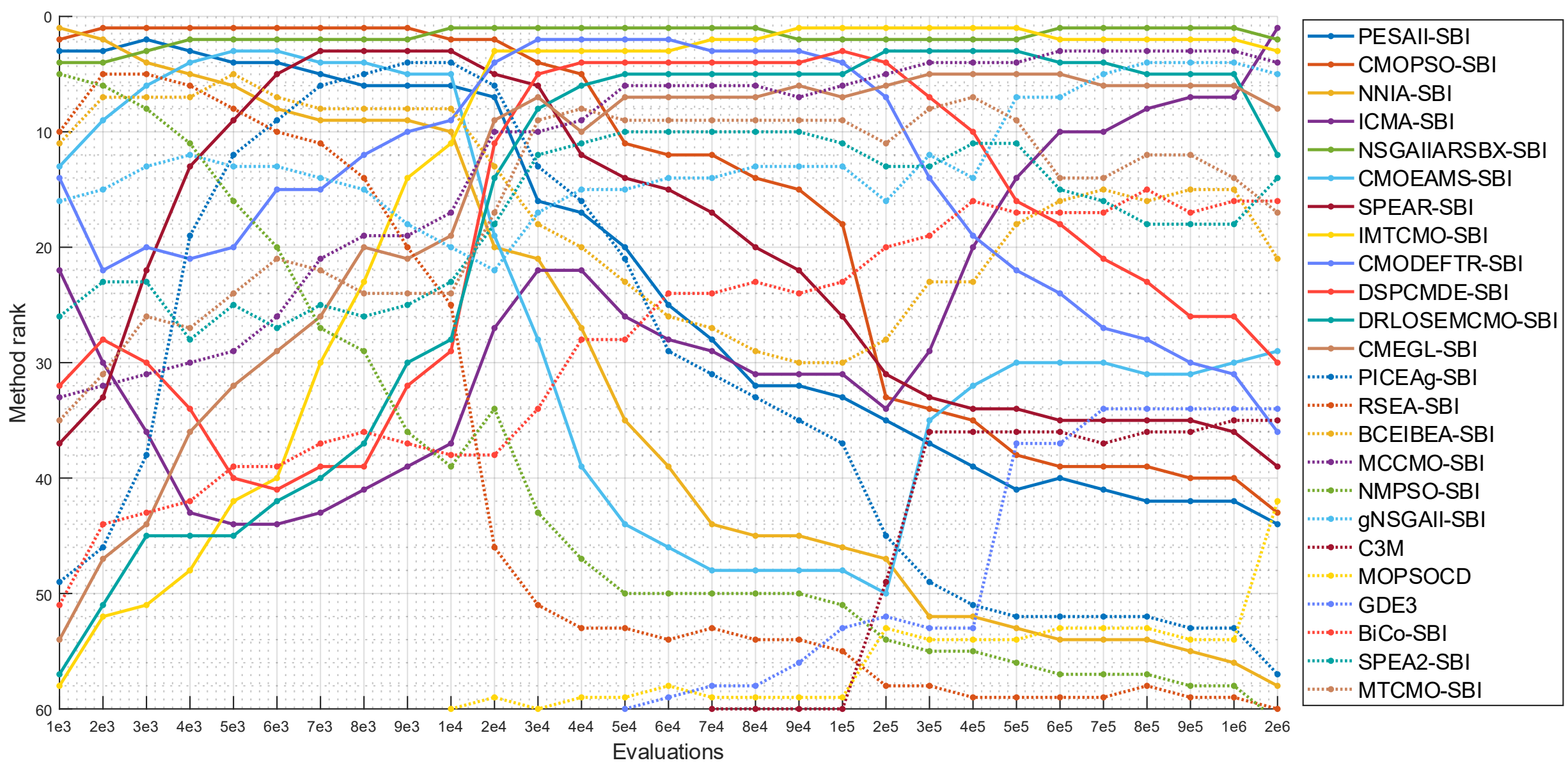

**Fig. 15.** Best methods based on their relative distance from the best-found HV values over the 24 problems for different computational budgets.

In Fig. 15 is shown the ranking (on the y axis the ranking could go up to 130 – the number of considered algorithm variants) of the selected methods based on the average relative distance, in which we can see patterns complementary to Fig. 12 and 13. By this assessment, NSGAIIARSBX-SBI is the overall best-performing method when considering all budgets and is in the top methods for any budget between 4e3 and 2e6. Then, there are specialized methods for lower budgets, such as NNIA-SBI, CMOPSO-SBI, SPEAR-SBI, which performance relative to other methods gradually decreases. There are also methods which are "specialized" for medium budgets (roughly between 1e4 and 1e5), like CMODEFTR-SBI and DSPCMDE-SBI, but which performance degrades afterwards. IMTCMO-SBI, DRLOSEMCMO-SBI and MCCMO-SBI also become one of the top methods only after the 1e4 budget, but are able to stay within the top 5 for the rest of the computational budgets. A similar thing could be said about ICMA-SBI, but it "comes online" only much later to become the method with the lowest achieved relative distance over all considered methods for the largest budget.

The relative distance values for all methods and computational budgets are shown in heatmaps on Fig. 16 (without SBI) and Fig. 17 (with SBI), where the methods are sorted by the last column (performance on the largest budget). What is perhaps a bit surprising is that the utilization of SBI changes the ranking of the methods quite drastically. None of the top three methods on the largest budget without SBI - GDE3, C3M, and MOPSOCD – was in the top 1/3 of methods with SBI. Of the top 10 methods with SBI, only IMTCMO was also in the top 10 methods without SBI, although it could get only to the relative distance of $8 \cdot 10^{-2}$. gNSGAII went from being top 5 in the SBI setting to being number 50 in the non-SBI one. These results raise an important point. The advantage of good initialization schemes (such as SBI) is not utilized equally by all methods. When such initialization schemes are not available, different methods should be preferred. This effect can most probably be attributed to the different levels of the exploration/exploitation capabilities of the different methods, but the analysis of this kind of behavior is beyond the scope of this paper.

### 5.5. *Effect of case-study complexity on algorithm performance*

Lastly, we turn our attention to **RQ4** (the link between the complexity of the CS and algorithm performance). Fig. 18 shows the boxplots of relative distance to the best-found HV for all methods, divided into different CS, budgets, and the utilization of SBI (the lowest values were capped at $10^{-8}$, because a "true 0" could not be used in a logarithmic plot). Here, we can see a few clear patterns. CS1 (5 subsystems, dimension 10) is the easiest problem, with half of the methods with SBI finding the same "best-possible" Pareto front solution. CS2 and CS3 (both 5 subsystems, dimension 10) are slightly more difficult than CS1, especially in larger evaluation budgets (meaning that there are less algorithms that can find "best-possible" solution sets). CS4 (10 subsystems, dimension 20) is more difficult still – especially when looking at the results of the non-SBI variants, none on which could find the "best-possible" solution set. Lastly, CS5 (10 subsystems, dimension 20) and CS6 (15 subsystems, dimension 30) are the most difficult ones where none of the methods (with or without SBI) could reliably get to the "best-possible" solution set.

The overall results show that the selected CS present a good range in difficulty and structure that manifests in significant differences in algorithmic performance of a wide range of methods over large computational budgets. This makes it the RAP a good candidate for becoming one of the standard real-world benchmark problems for multi-objective optimization algorithms.

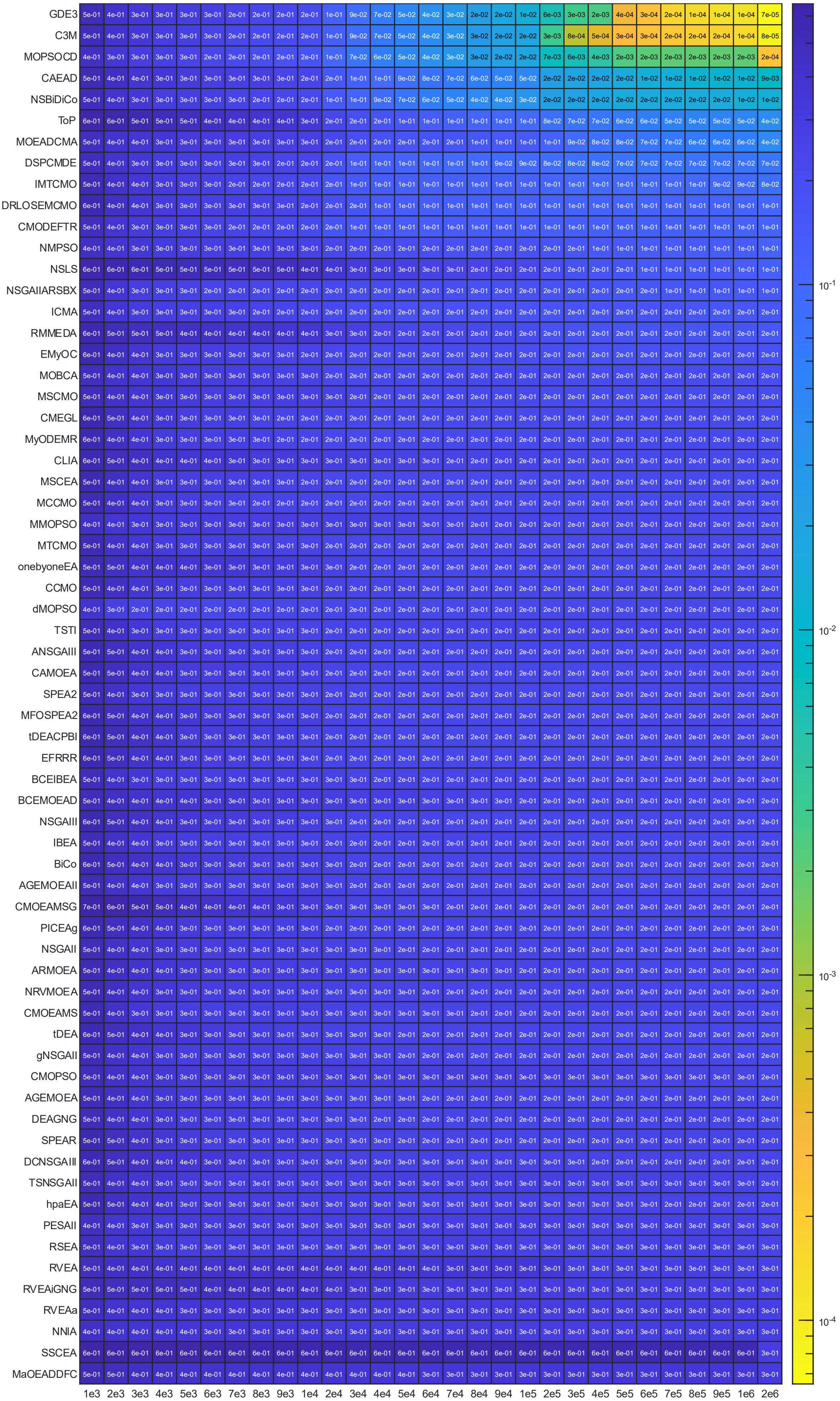

**Fig. 16.** Relative distance from the best-found HV values over the 24 problems. Sorted by the last column. Methods without SBI.

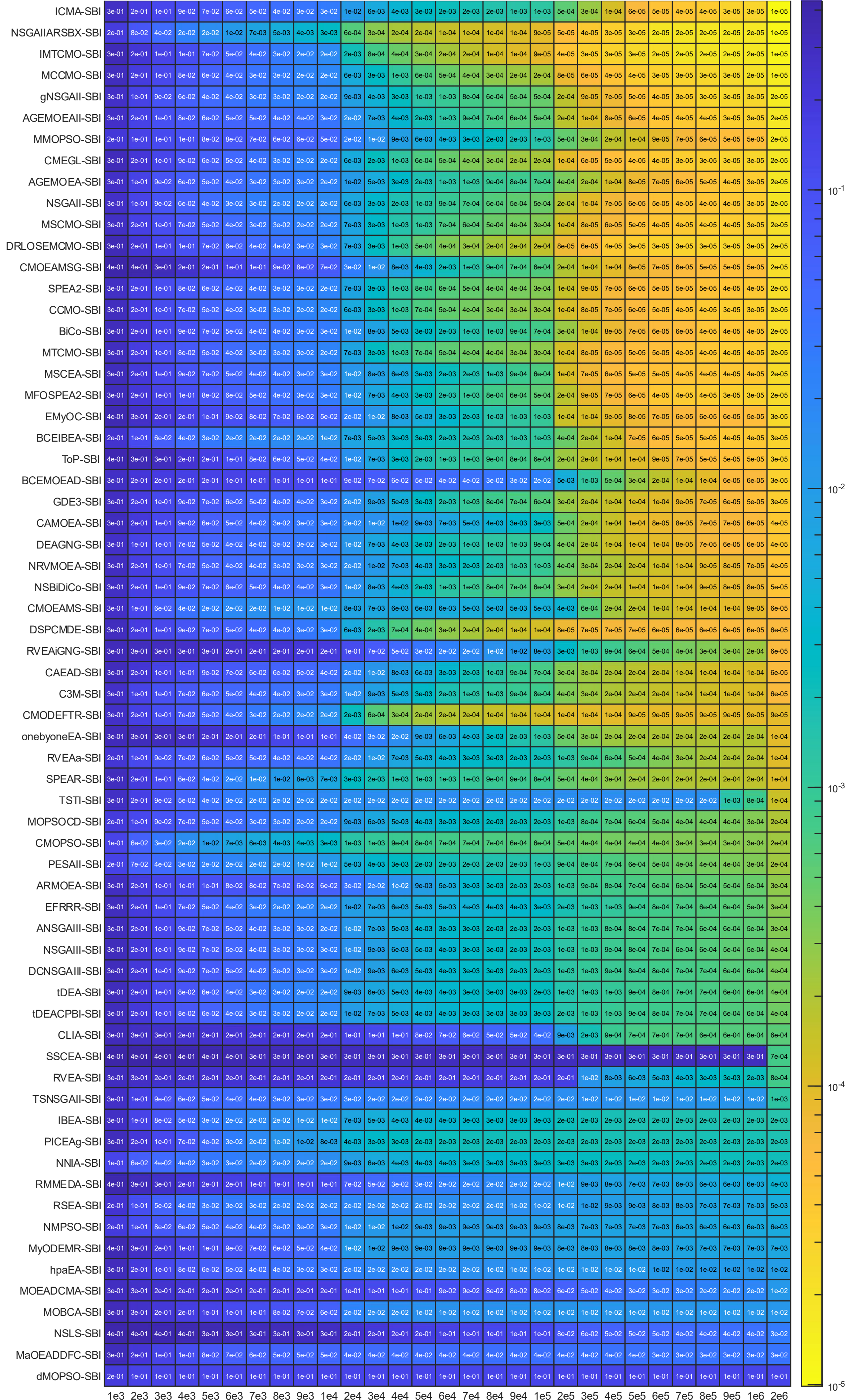

**Fig. 17.** Relative distance from the best-found HV values over the 24 problems. Sorted by the last column. Methods with SBI.

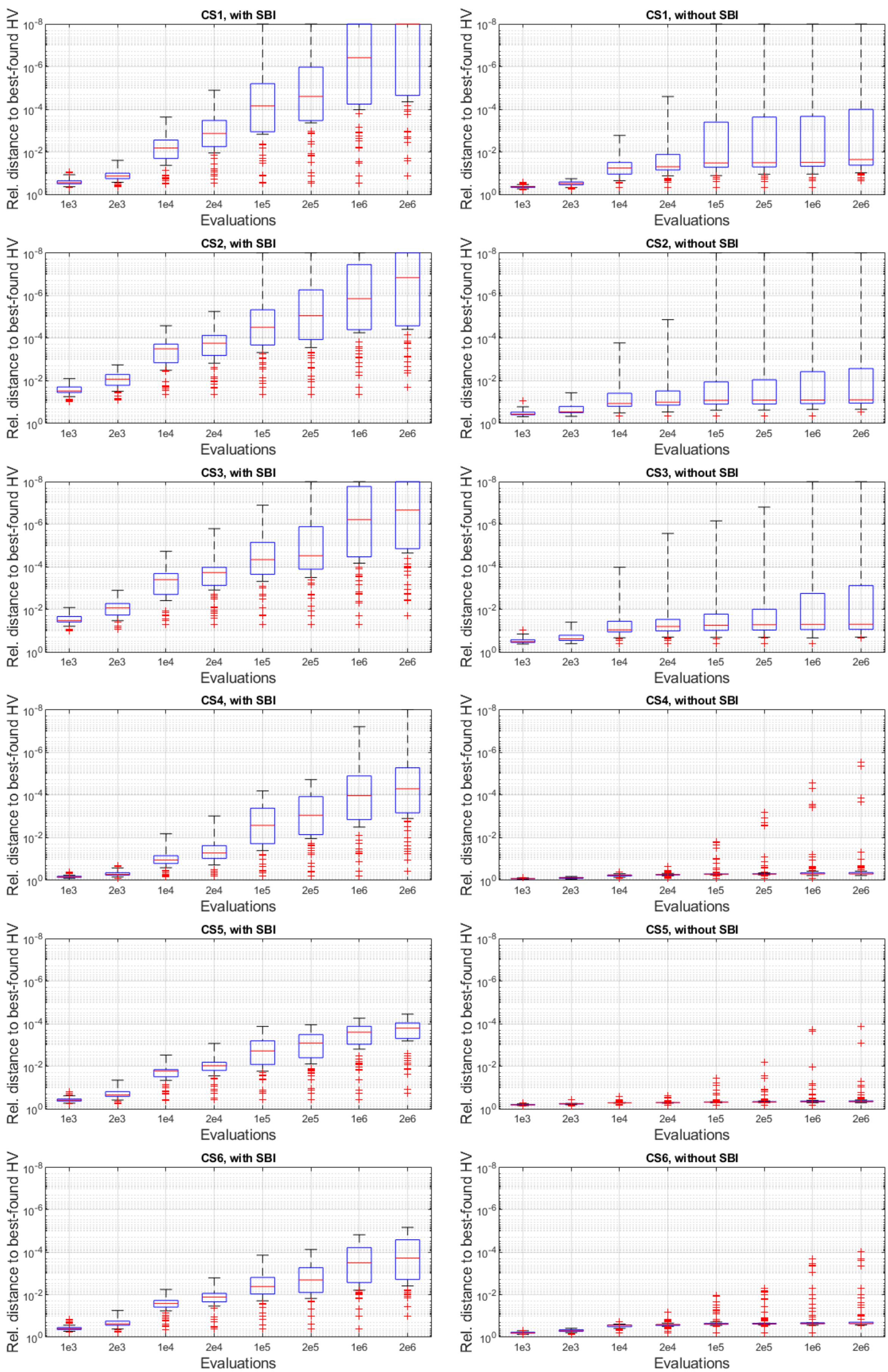

**Fig. 18.** Boxplots of relative distance from the best-found HV values (average over 10 runs) for all methods divided into the different CS and by the utilization of SBI.

## 6. Conclusions and future work

A large-scale comparative analysis of hybrid and state-of-the-art multi-objective metaheuristics was conducted for a bi-objective RAP in repairable systems, formulated as the joint minimization of cost and maximization of system availability. Sixty-five algorithms were evaluated across six case studies of increasing structural and dimensional complexity and four weight-constraint settings. Two initialization regimes were considered (with and without SBI), and computational budgets of up to $2 \times 10^6$ function evaluations were employed, with repeated runs to enable statistically supported comparisons. The setting uncovered a multitude of statistically significant differences in the performance of the different methods and provide guidelines for efficient selection for solving the multiobjective RAP.

From a reliability-engineering standpoint, four redundancy strategies were examined. A clear dominance of hot standby was observed under tight weight constraints, whereas the mixed strategy was favored when larger numbers of redundant components could be allocated to subsystems, irrespective of the overall technical-system complexity. Cold and warm standby occurred only marginally in the full solution populations and were effectively absent from the aggregated Pareto fronts, indicating that these strategies are not competitive under the considered model and parameterization.

From an algorithmic perspective, the results indicate that “best” performance is inherently budget-dependent; consequently, benchmarking outcomes cannot be faithfully summarized by a single global ranking without conditioning on the available evaluation budget. Within the set of ten algorithm groups classified according to their operating principles, the most effective approaches were identified. In particular, NSGAIIARSBX-SBI exhibited consistently strong performance across a wide range of computational budgets, which supports its characterization as a robust general-purpose choice for medium and large evaluation regimes. For small budgets, NNIA-SBI and CMOPSO-SBI were found to be the most competitive.

The use of SBI for initial population generation was shown to confer a substantial advantage in terms of HV across the tested budgets, extending prior evidence previously reported primarily in the context of NSGA-II [55]. SBI consistently produced higher HV values, and in selected case studies the SBI-generated population alone achieved HV levels close to the best-found reference, effectively shifting the search into a later stage of convergence from the outset. Only a limited number of algorithms using conventional random initialization attained performance comparable to the top third of SBI-based methods, namely GDE3, C3M, and MOPSOCD. Overall, the findings provide strong support for employing SBI in binary-encoded RAP settings and motivate further investigation of problem-informed initialization mechanisms that can yield systematic performance gains.

Increasing system complexity, reflected by a larger number of subsystems to be optimized with respect to the number of spares and redundancy-strategy selection, was found to markedly affect algorithm behavior by slowing progress toward the global Pareto front and increasing the computational effort required to achieve high-quality convergence. For small-scale systems, rapid convergence of the Pareto front was typically observed at approximately $10^4$ evaluations, whereas for large-scale systems at least $10^5$ evaluations were required to attain a comparably high level of convergence. From a practical standpoint, accurate estimation of the computational budget required to obtain Pareto fronts close to global optima is therefore essential. In summary, these observations suggest that the selected case studies span a meaningful gradient of difficulty and structure and may serve as a valuable real-world benchmark suite for multi-objective optimization research in redundancy allocation.

Future research directions in this area may focus on several important aspects. First, comparative benchmarking studies of systems exhibiting heterogeneous reliability characteristics, such as time-varying failure and repair rates, may provide valuable insights into the behavior of hybrid metaheuristics under more dynamic operating conditions. Such analyses could also encompass alternative system configurations and investigate dependencies between failures and repairs in scenarios with limited maintenance resources, thereby reflecting more realistic challenges encountered in practical applications. From an algorithmic standpoint, substantial potential exists to refine initial population generation methods in order to improve algorithmic efficiency beyond currently available approaches. The development of more effective initialization techniques may result in faster convergence and more accurate solutions, thereby offering a competitive advantage when addressing complex optimization problems. Moreover, future work could include benchmarking studies for multi-objective RAP formulations aimed at maximizing system reliability or availability while simultaneously minimizing cost, weight, and volume.

### Acknowledgements

This work was supported by the project GACR No. 24-12474S “Benchmarking derivative-free global optimization methods” and by the project IGA BUT No. FSI-S-23-8394 “Artificial intelligence methods in engineering tasks” and by the project “Mechanical Engineering of Biological and Bio-inspired Systems”, funded as project No. CZ.02.01.01/00/22_008/0004634 by Programme Johannes Amos Commenius, call Excellent Research. Computational resources were provided by the e-INFRA CZ project (ID:90254), supported by the Ministry of Education, Youth and Sports of the Czech Republic.

### References

[1] S.M. Mousavi, N. Alikar, M. Tavana, D. Di Caprio, An improved particle swarm optimization model for solving homogeneous discounted series-parallel redundancy allocation problems, J Intell Manuf 30 (2019) 1175–1194. https://doi.org/10.1007/s10845-017-1311-9.

[2] H.A. Khorshidi, I. Gunawan, M.Y. Ibrahim, A value-driven approach for optimizing reliability-redundancy allocation problem in multi-state weighted k-out-of-n system, Journal of Manufacturing Systems 40 (2016) 54–62. https://doi.org/10.1016/j.jmsy.2016.06.002.

[3] C. Cheng, J. Yang, L. Li, Reliability evaluation of a k-out-of-n(G)-subsystem based multi-state phased mission system with common bus performance sharing subjected to common cause failures, Reliability Engineering & System Safety 216 (2021) 108003. https://doi.org/10.1016/j.ress.2021.108003.

[4] K. Hamdan, M. Tavangar, M. Asadi, Optimal preventive maintenance for repairable weighted k-out-of-n systems, Reliability Engineering & System Safety 205 (2021) 107267. https://doi.org/10.1016/j.ress.2020.107267.

[5] I. Betkier, Estimating travel time in transport network with a combined multi-attributed graph convolutional neural network and multilayer perceptron model, Engineering Applications of Artificial Intelligence 142 (2025) 109898. https://doi.org/10.1016/j.engappai.2024.109898.

[6] Md.A.M. Chowdury, R. Nath, A.K. Shukla, A. Rauniyar, P.K. Muhuri, Multi-task optimization in reliability redundancy allocation problem: A multifactorial evolutionary-based approach, Reliability Engineering & System Safety 244 (2024) 109807. https://doi.org/10.1016/j.ress.2023.109807.

[7] T.-J. Hsieh, Component mixing with a cold standby strategy for the redundancy allocation problem, Reliability Engineering & System Safety 206 (2021) 107290. https://doi.org/10.1016/j.ress.2020.107290.

[8] G. Levitin, L. Xing, Y. Dai, Optimal operation and maintenance scheduling in generalized repairable m-out-of-n standby systems with common shocks, Reliability Engineering & System Safety 260 (2025) 110967. https://doi.org/10.1016/j.ress.2025.110967.

[9] G. Levitin, L. Xing, H.Z. Huang, Dynamic availability and performance deficiency of common bus systems with imperfectly repairable components, Reliability Engineering & System Safety 189 (2019) 58–66. https://doi.org/10.1016/j.ress.2019.04.007.

[10] M. Oszczypała, J. Konwerski, J. Ziółkowski, J. Małachowski, Reliability analysis and redundancy optimization of k-out-of-n systems with random variable k using continuous time Markov chain and Monte Carlo simulation, Reliability Engineering & System Safety 242 (2024) 109780. https://doi.org/10.1016/j.ress.2023.109780.

[11] A. Peiravi, M. Nourelfath, M.K. Zanjani, Redundancy strategies assessment and optimization of k-out-of-n systems based on Markov chains and genetic algorithms, Reliability Engineering & System Safety 221 (2022) 108277. https://doi.org/10.1016/j.ress.2021.108277.

[12] A. Peiravi, M. Nourelfath, M.K. Zanjani, Universal redundancy strategy for system reliability optimization, Reliability Engineering & System Safety 225 (2022) 108576. https://doi.org/10.1016/j.ress.2022.108576.

[13] S. Zhao, Y. Wei, Y. Cheng, Y. Li, A state-specific joint size, maintenance, and inventory policy for a k-out-of-n load-sharing system subject to self-announcing failures, Reliability Engineering & System Safety 257 (2025) 110855. https://doi.org/10.1016/j.ress.2025.110855.

[14] J. Zhong, H. Zhang, Q. Miao, Enhancing aircraft reliability with information redundancy: A sensor-modal fusion approach leveraging deep learning, Reliability Engineering & System Safety 261 (2025) 111068. https://doi.org/10.1016/j.ress.2025.111068.

[15] G. Chen, Y. Hu, C. Wang, Z. Wu, W. Rong, Q. Guan, Efficient reliability analysis of generalized k -out-of- n phased-mission systems, Reliability Engineering & System Safety 253 (2025) 110581. https://doi.org/10.1016/j.ress.2024.110581.

[16] T. Xiahou, Y.-X. Zheng, Y. Liu, H. Chen, Reliability modeling of modular k-out-of-n systems with functional dependency: A case study of radar transmitter systems, Reliability Engineering & System Safety 233 (2023) 109120. https://doi.org/10.1016/j.ress.2023.109120.

[17] M. Oszczypała, J. Ziółkowski, J. Małachowski, Redundancy allocation problem in repairable k-out-of-n systems with cold, warm, and hot standby: A genetic algorithm for availability optimization, Applied Soft Computing 165 (2024) 112041. https://doi.org/10.1016/j.asoc.2024.112041.

[18] A. Singh, S.B. Singh, Dynamic reliability and sensitivity analysis of weighted (k, r)-out-of-n cold standby system with multi-performance multi-state components, Reliability Engineering & System Safety 262 (2025) 111221. https://doi.org/10.1016/j.ress.2025.111221.

[19] J. Li, D. Wang, H. Yang, M. Liu, S. Si, An exact algorithm for RAP with k-out-of-n subsystems and heterogeneous components under mixed and K-mixed redundancy strategies, Advanced Engineering Informatics 65 (2025) 103163. https://doi.org/10.1016/j.aei.2025.103163.

[20] D. Wang, M. Liu, H. Yang, S. Si, A novel importance measure considering multi-constraints for RAP optimization of 1-out-of-n subsystems with mixed redundancy strategy, Reliability Engineering & System Safety 252 (2024) 110441. https://doi.org/10.1016/j.ress.2024.110441.

[21] L. Stripinis, J. Kůdela, R. Paulavičius, Benchmarking Derivative-Free Global Optimization Algorithms Under Limited Dimensions and Large Evaluation Budgets, IEEE Trans. Evol. Computat. 29 (2025) 187–204. https://doi.org/10.1109/TEVC.2024.3379756.

[22] J. Kůdela, M. Zálešák, P. Charvát, L. Klimeš, T. Mauder, Assessment of the performance of metaheuristic methods used for the inverse identification of effective heat capacity of phase change materials, Expert Systems with Applications 238 (2024) 122373. https://doi.org/10.1016/j.eswa.2023.122373.

[23] J. Pluskal, B. Zach, J. Kůdela, R. Šomplák, M. Šyc, Membrane-based carbon capture for waste-to-energy: Process performance, impact, and time-efficient optimization, Energy 310 (2024) 133229. https://doi.org/10.1016/j.energy.2024.133229.

[24] M. Ali Shehadeh, J. Kůdela, Benchmarking global optimization techniques for unmanned aerial vehicle path planning, Expert Systems with Applications 293 (2025) 128645. https://doi.org/10.1016/j.eswa.2025.128645.

[25] T. Bartz-Beielstein, C. Doerr, D. van den Berg, J. Bossek, S. Chandrasekaran, T. Eftimov, A. Fischbach, P. Kerschke, W. La Cava, M. Lopez-Ibanez, K.M. Malan, J.H. Moore, B. Naujoks, P. Orzechowski, V. Volz, M. Wagner, T. Weise, Benchmarking in Optimization: Best Practice and Open Issues, (2020). https://doi.org/10.48550/ARXIV.2007.03488.

[26] L. Stripinis, J. Kůdela, R. Paulavičius, Two Novel Instance Selection Methods Combining Algorithm Performance and Landscape Analysis: A Comparative Study in Continuous Optimization, IEEE Trans. Cybern. (2025) 1–14. https://doi.org/10.1109/TCYB.2025.3625095.

[27] B. Doerr, Z. Qu, Runtime Analysis for the NSGA-II: Provable Speed-Ups from Crossover, AAAI 37 (2023) 12399–12407. https://doi.org/10.1609/aaai.v37i10.26461.

[28] S. Wietheger, B. Doerr, A Mathematical Runtime Analysis of the Non-dominated Sorting Genetic Algorithm III (NSGA-III), in: Proceedings of the Genetic and Evolutionary Computation Conference Companion, ACM, Melbourne VIC Australia, 2024: pp. 63–64. https://doi.org/10.1145/3638530.3664062.

[29] W. Zheng, B. Doerr, Runtime Analysis for the NSGA-II: Proving, Quantifying, and Explaining the Inefficiency for Many Objectives, IEEE Trans. Evol. Computat. 28 (2024) 1442–1454. https://doi.org/10.1109/TEVC.2023.3320278.

[30] A. Auger, D. Brockhoff, J. Cork, T. Tušar, On the Pareto Set and Front of Multiobjective Spherical Functions with Convex Constraints, in: Proceedings of the Genetic and Evolutionary Computation Conference, ACM, NH Malaga Hotel Malaga Spain, 2025: pp. 527–535. https://doi.org/10.1145/3712256.3726432.

[31] K. Deb, L. Thiele, M. Laumanns, E. Zitzler, Scalable Test Problems for Evolutionary Multiobjective Optimization, in: A. Abraham, L. Jain, R. Goldberg (Eds.), Evolutionary Multiobjective Optimization, Springer-Verlag, London, 2005: pp. 105–145. https://doi.org/10.1007/1-84628-137-7_6.

[32] S. Huband, P. Hingston, L. Barone, L. While, A review of multiobjective test problems and a scalable test problem toolkit, IEEE Trans. Evol. Computat. 10 (2006) 477–506. https://doi.org/10.1109/TEVC.2005.861417.

[33] T. Tusar, On using real-world problems for benchmarking multiobjective optimization algorithms, in: Proceedings of the International Conference on High-Performance Optimization in Industry, International Multiconference Information Society, Jožef Stefan Institute, 2018: pp. 7–10. https://dis.ijs.si/tea/Publications/Tusar18Using.pdf.

[34] D. Brockhoff, A. Auger, N. Hansen, T. Tušar, Using Well-Understood Single-Objective Functions in Multiobjective Black-Box Optimization Test Suites, Evolutionary Computation (2022) 165–193. https://doi.org/10.1162/evco_a_00298.

[35] N. Hansen, A. Auger, R. Ros, O. Mersmann, T. Tušar, D. Brockhoff, COCO: a platform for comparing continuous optimizers in a black-box setting, Optimization Methods and Software 36 (2021) 114–144. https://doi.org/10.1080/10556788.2020.1808977.

[36] P. Krömer, V. Uher, T. Tušar, B. Filipič, On the Latent Structure of the bbob-biobj Test Suite, in: S. Smith, J. Correia, C. Cintrano (Eds.), Applications of Evolutionary Computation, Springer Nature Switzerland, Cham, 2024: pp. 326–341. https://doi.org/10.1007/978-3-031-56855-8_20.

[37] A.V. Kononova, N. van Stein, O. Mersmann, T. Bäck, T. Bartz-Beielstein, T. Glasmachers, M. Hellwig, S. Krey, J. Kůdela, B. Naujoks, L. Papenmeier, E. Raponi, Q. Renau, J. Rook, L. Schäpermeier, D. Vermetten, D. Zaharie, Benchmarking that Matters: Rethinking Benchmarking for Practical Impact, (2025). https://doi.org/10.48550/ARXIV.2511.12264.

[38] J. Kudela, A critical problem in benchmarking and analysis of evolutionary computation methods, Nat Mach Intell 4 (2022) 1238–1245. https://doi.org/10.1038/s42256-022-00579-0.

[39] E.J. Hughes, Radar Waveform Optimisation as a Many-Objective Application Benchmark, in: S. Obayashi, K. Deb, C. Poloni, T. Hiroyasu, T. Murata (Eds.), Evolutionary Multi-Criterion Optimization, Springer Berlin Heidelberg, Berlin, Heidelberg, 2007: pp. 700–714. https://doi.org/10.1007/978-3-540-70928-2_53.

[40] P.M. Reed, D. Hadka, J.D. Herman, J.R. Kasprzyk, J.B. Kollat, Evolutionary multiobjective optimization in water resources: The past, present, and future, Advances in Water Resources 51 (2013) 438–456. https://doi.org/10.1016/j.advwatres.2012.01.005.

[41] T. Kohira, H. Kemmotsu, O. Akira, T. Tatsukawa, Proposal of benchmark problem based on real-world car structure design optimization, in: Proceedings of the Genetic and Evolutionary Computation Conference Companion, ACM, Kyoto Japan, 2018: pp. 183–184. https://doi.org/10.1145/3205651.3205702.

[42] J. Li, B. Xin, P.M. Pardalos, J. Chen, Solving bi-objective uncertain stochastic resource allocation problems by the CVaR-based risk measure and decomposition-based multi-objective evolutionary algorithms, Ann Oper Res 296 (2021) 639–666. https://doi.org/10.1007/s10479-019-03435-4.

[43] W.-C. Yeh, Y.-Z. Su, X.-Z. Gao, C.-F. Hu, J. Wang, C.-L. Huang, Simplified swarm optimization for bi-objection active reliability redundancy allocation problems, Applied Soft Computing 106 (2021) 107321. https://doi.org/10.1016/j.asoc.2021.107321.

[44] F. Hadinejad, M. Amiri, Redundancy allocation problem in repairable systems with variegated components: a simulation-based optimization approach, Journal of Statistical Computation and Simulation 94 (2024) 3301–3318. https://doi.org/10.1080/00949655.2024.2383722.

[45] F. Kayedpour, M. Amiri, M. Rafizadeh, A.S. Nia, M. Sharifi, A Markov chain-based genetic algorithm for solving a redundancy allocation problem for a system with repairable warm standby components, Proceedings of the Institution of Mechanical Engineers, Part O: Journal of Risk and Reliability 238 (2024) 853–872. https://doi.org/10.1177/1748006X231164848.

[46] A. Zaretalab, M. Sharifi, P.P. Guilani, S. Taghipour, S.T.A. Niaki, A multi-objective model for optimizing the redundancy allocation, component supplier selection, and reliable activities for multi-state systems, Reliability Engineering & System Safety 222 (2022) 108394. https://doi.org/10.1016/j.ress.2022.108394.

[47] A. Chambari, P. Azimi, A.A. Najafi, A bi-objective simulation-based optimization algorithm for redundancy allocation problem in series-parallel systems, Expert Systems with Applications 173 (2021) 114745. https://doi.org/10.1016/j.eswa.2021.114745.

[48] M.N. Juybari, P.P. Guilani, M.A. Ardakan, Bi-objective sequence optimization in reliability problems with a matrix-analytic approach, Ann Oper Res 312 (2022) 275–304. https://doi.org/10.1007/s10479-021-04039-7.

[49] M. Yaghtin, Y. Javid, M.A. Ardakan, Multi-objective optimization in the design of load sharing systems with mixed redundancy strategies under random shocks, Journal of Computational Science 85 (2025) 102495. https://doi.org/10.1016/j.jocs.2024.102495.

[50] G.S. Mahapatra, B. Maneckshaw, K. Barker, Multi-objective reliability redundancy allocation using MOPSO under hesitant fuzziness, Expert Systems with Applications 198 (2022) 116696. https://doi.org/10.1016/j.eswa.2022.116696.

[51] S. De, P. Roy, S. Roy, A.B. Chowdhury, Optimization of time-dependent fuzzy multi-objective reliability redundancy allocation problem for n-stage series–parallel system, Innovations Syst Softw Eng 21 (2025) 801–811. https://doi.org/10.1007/s11334-023-00539-w.

[52] S. De, Optimization of time-dependent MORRAP for series–parallel system using improved NSGA-II in interval environment, Innovations Syst Softw Eng 21 (2025) 1025–1039. https://doi.org/10.1007/s11334-024-00588-9.

[53] S. De, P. Rakshit, A.B. Chowdhury, Optimization of time based fuzzy multi-objective reliability redundancy allocation problem for x j − o u t − o f − m system using tuning and neighborhood based fuzzy MOPSO algorithm, Applied Soft Computing 149 (2023) 110998. https://doi.org/10.1016/j.asoc.2023.110998.

[54] Y. Xu, D. Pi, S. Yang, Y. Chen, S. Qin, E. Zio, An Angle-Based Bi-Objective Optimization Algorithm for Redundancy Allocation in Presence of Interval Uncertainty, IEEE Trans. Automat. Sci. Eng. 20 (2023) 271–284. https://doi.org/10.1109/TASE.2022.3148459.

[55] M. Oszczypała, Bi-objective redundancy allocation problem in systems with mixed strategy: NSGA-II with a novel initialization, Reliability Engineering & System Safety 263 (2025) 111279. https://doi.org/10.1016/j.ress.2025.111279.

[56] R. Matousek, L. Dobrovsky, J. Kudela, How to start a heuristic? Utilizing lower bounds for solving the quadratic assignment problem, 10.5267/j.Ijiec 13 (2022) 151–164. https://doi.org/10.5267/j.ijiec.2021.12.003.

[57] Y. Liu, G. Wang, P. Liu, A condition-based maintenance policy with non-periodic inspection for k-out-of-n: G systems, Reliability Engineering & System Safety 241 (2024) 109640. https://doi.org/10.1016/j.ress.2023.109640.

[58] S. Choudhary, M. Ram, N. Goyal, Reliability optimization of non-linear RRAP with cold standby through HPSOTLBO, Computers & Industrial Engineering 203 (2025) 111045. https://doi.org/10.1016/j.cie.2025.111045.

[59] S. Faghih-Roohi, M. Xie, K.M. Ng, R.C.M. Yam, Dynamic availability assessment and optimal component design of multi-state weighted k-out-of-n systems, Reliability Engineering & System Safety 123 (2014) 57–62. https://doi.org/10.1016/j.ress.2013.10.002.

[60] L. Cui, W. Jiang, M. Wang, Reliability Analysis for Warm Standby Systems with Two Different Spare Areas, in: Q.Q. Zhao, I.H. Chung, J. Zheng, J. Kim (Eds.), Reliability Analysis and Maintenance Optimization of Complex Systems, Springer Nature Switzerland, Cham, 2025: pp. 45–63. https://doi.org/10.1007/978-3-031-70288-4_4.

[61] I.S. Triantafyllou, Consecutive-type coherent systems with cold standby redundancy at the system level: Advances and applications, in: Reliability Assessment and Optimization of Complex Systems, Elsevier, 2025: pp. 23–35. https://doi.org/10.1016/B978-0-443-29112-8.00013-X.

[62] J.Y. Kim, L. Mastropasqua, A. Saeedmanesh, J. Brouwer, Development of thermal control strategies for solid oxide electrolysis cell systems under dynamic operating conditions - Hot-standby and cold-start scenarios, Energy 317 (2025) 134679. https://doi.org/10.1016/j.energy.2025.134679.

[63] V.P. Singh, M. Jain, R. Sharma, N-policy for redundant machining system with double retrial orbits using soft computing techniques, Mathematics and Computers in Simulation 237 (2025) 42–69. https://doi.org/10.1016/j.matcom.2025.04.025.

[64] Y. Tian, R. Cheng, X. Zhang, Y. Jin, PlatEMO: A MATLAB Platform for Evolutionary Multi-Objective Optimization [Educational Forum], IEEE Comput. Intell. Mag. 12 (2017) 73–87. https://doi.org/10.1109/MCI.2017.2742868.

[65] D. Ibehej, J. Kudela, Benchmarking Seven Multi-objective Optimization Methods from the PlatEMO Platform on the bbob-biobj Test Suite, in: Proceedings of the Genetic and Evolutionary Computation Conference Companion, ACM, NH Malaga Hotel Malaga Spain, 2025: pp. 1883–1890. https://doi.org/10.1145/3712255.3734349.

[66] H. Jain, K. Deb, An Evolutionary Many-Objective Optimization Algorithm Using Reference-Point Based Nondominated Sorting Approach, Part II: Handling Constraints and Extending to an Adaptive Approach, IEEE Trans. Evol. Computat. 18 (2014) 602–622. https://doi.org/10.1109/TEVC.2013.2281534.

[67] Y. Zhang, Y. Tian, H. Jiang, X. Zhang, Y. Jin, Design and analysis of helper-problem-assisted evolutionary algorithm for constrained multiobjective optimization, Information Sciences 648 (2023) 119547. https://doi.org/10.1016/j.ins.2023.119547.

[68] A. Panichella, An adaptive evolutionary algorithm based on non-euclidean geometry for many-objective optimization, in: Proceedings of the Genetic and Evolutionary Computation Conference, ACM, Prague Czech Republic, 2019: pp. 595–603. https://doi.org/10.1145/3321707.3321839.

[69] H. Ma, H. Wei, Y. Tian, R. Cheng, X. Zhang, A multi-stage evolutionary algorithm for multi-objective optimization with complex constraints, Information Sciences 560 (2021) 68–91. https://doi.org/10.1016/j.ins.2021.01.029.

[70] A. Panichella, An improved Pareto front modeling algorithm for large-scale many-objective optimization, in: Proceedings of the Genetic and Evolutionary Computation Conference, ACM, Boston Massachusetts, 2022: pp. 565–573. https://doi.org/10.1145/3512290.3528732.

[71] K. Qiao, K. Yu, B. Qu, J. Liang, H. Song, C. Yue, H. Lin, K.C. Tan, Dynamic Auxiliary Task-Based Evolutionary Multitasking for Constrained Multiobjective Optimization, IEEE Trans. Evol. Computat. 27 (2023) 642–656. https://doi.org/10.1109/TEVC.2022.3175065.

[72] Y. Tian, R. Cheng, X. Zhang, F. Cheng, Y. Jin, An Indicator-Based Multiobjective Evolutionary Algorithm With Reference Point Adaptation for Better Versatility, IEEE Trans. Evol. Computat. 22 (2018) 609–622. https://doi.org/10.1109/TEVC.2017.2749619.

[73] Jixiang Cheng, G.G. Yen, Gexiang Zhang, A Many-Objective Evolutionary Algorithm With Enhanced Mating and Environmental Selections, IEEE Trans. Evol. Computat. 19 (2015) 592–605. https://doi.org/10.1109/TEVC.2015.2424921.

[74] M. Li, S. Yang, X. Liu, Pareto or Non-Pareto: Bi-Criterion Evolution in Multiobjective Optimization, IEEE Trans. Evol. Computat. 20 (2016) 645–665. https://doi.org/10.1109/TEVC.2015.2504730.

[75] R. Denysiuk, L. Costa, I. Espírito Santo, Many-objective optimization using differential evolution with variable-wise mutation restriction, in: Proceedings of the 15th Annual Conference on Genetic and Evolutionary Computation, ACM, Amsterdam The Netherlands, 2013: pp. 591–598. https://doi.org/10.1145/2463372.2463445.

[76] Q. Lin, S. Liu, Q. Zhu, C. Tang, R. Song, J. Chen, C.A.C. Coello, K.-C. Wong, J. Zhang, Particle Swarm Optimization With a Balanceable Fitness Estimation for Many-Objective Optimization Problems, IEEE Trans. Evol. Computat. 22 (2018) 32–46. https://doi.org/10.1109/TEVC.2016.2631279.

[77] Z.-Z. Liu, B.-C. Wang, K. Tang, Handling Constrained Multiobjective Optimization Problems via Bidirectional Coevolution, IEEE Trans. Cybern. 52 (2022) 10163–10176. https://doi.org/10.1109/TCYB.2021.3056176.

[78] M. Gong, L. Jiao, H. Du, L. Bo, Multiobjective Immune Algorithm with Nondominated Neighbor-Based Selection, Evolutionary Computation 16 (2008) 225–255. https://doi.org/10.1162/evco.2008.16.2.225.

[79] R. Sun, J. Zou, Y. Liu, S. Yang, J. Zheng, A Multistage Algorithm for Solving Multiobjective Optimization Problems With Multiconstraints, IEEE Trans. Evol. Computat. 27 (2023) 1207–1219. https://doi.org/10.1109/TEVC.2022.3224600.

[80] Y. Hua, Q. Liu, K. Hao, Adaptive normal vector guided evolutionary multi- and many-objective optimization, Complex Intell. Syst. 10 (2024) 3709–3726. https://doi.org/10.1007/s40747-024-01353-y.

[81] Y. Hua, Y. Jin, K. Hao, A Clustering-Based Adaptive Evolutionary Algorithm for Multiobjective Optimization With Irregular Pareto Fronts, IEEE Trans. Cybern. 49 (2019) 2758–2770. https://doi.org/10.1109/TCYB.2018.2834466.

[82] C.S.R. Mendes, A.F.R. Araújo, L.R.C. Farias, Non-Dominated Sorting Bidirectional Differential Coevolution, in: 2023 IEEE International Conference on Systems, Man, and Cybernetics (SMC), IEEE, Honolulu, Oahu, HI, USA, 2023: pp. 1709–1714. https://doi.org/10.1109/SMC53992.2023.10394195.

[83] J. Zou, R. Sun, S. Yang, J. Zheng, A dual-population algorithm based on alternative evolution and degeneration for solving constrained multi-objective optimization problems, Information Sciences 579 (2021) 89–102. https://doi.org/10.1016/j.ins.2021.07.078.

[84] K. Deb, A. Pratap, S. Agarwal, T. Meyarivan, A fast and elitist multiobjective genetic algorithm: NSGA-II, IEEE Trans. Evol. Computat. 6 (2002) 182–197. https://doi.org/10.1109/4235.996017.

[85] Y. Tian, T. Zhang, J. Xiao, X. Zhang, Y. Jin, A Coevolutionary Framework for Constrained Multiobjective Optimization Problems, IEEE Trans. Evol. Computat. 25 (2021) 102–116. https://doi.org/10.1109/TEVC.2020.3004012.

[86] L. Pan, W. Xu, L. Li, C. He, R. Cheng, Adaptive simulated binary crossover for rotated multi-objective optimization, Swarm and Evolutionary Computation 60 (2021) 100759. https://doi.org/10.1016/j.swevo.2020.100759.

[87] H. Ge, M. Zhao, L. Sun, Z. Wang, G. Tan, Q. Zhang, C.L.P. Chen, A Many-Objective Evolutionary Algorithm With Two Interacting Processes: Cascade Clustering and Reference Point Incremental Learning, IEEE Trans. Evol. Computat. 23 (2019) 572–586. https://doi.org/10.1109/TEVC.2018.2874465.

[88] K. Deb, H. Jain, An Evolutionary Many-Objective Optimization Algorithm Using Reference-Point-Based Nondominated Sorting Approach, Part I: Solving Problems With Box Constraints, IEEE Trans. Evol. Computat. 18 (2014) 577–601. https://doi.org/10.1109/TEVC.2013.2281535.

[89] K. Qiao, J. Liang, Z. Liu, K. Yu, C. Yue, B. Qu, Evolutionary Multitasking with Global and Local Auxiliary Tasks for Constrained Multi-Objective Optimization, IEEE/CAA J. Autom. Sinica 10 (2023) 1951–1964. https://doi.org/10.1109/JAS.2023.123336.

[90] Bili Chen, Wenhua Zeng, Yangbin Lin, Defu Zhang, A New Local Search-Based Multiobjective Optimization Algorithm, IEEE Trans. Evol. Computat. 19 (2015) 50–73. https://doi.org/10.1109/TEVC.2014.2301794.

[91] Z. Zeng, X. Zhang, Z. Hong, A constrained multiobjective differential evolution algorithm based on the fusion of two rankings, Information Sciences 647 (2023) 119572. https://doi.org/10.1016/j.ins.2023.119572.

[92] D.W. Corne, N.R. Jerram, J.D. Knowles, M.J. Oates, PESA-II: region-based selection in evolutionary multiobjective optimization, in: Proceedings of the 3rd Annual Conference on Genetic and Evolutionary Computation, Morgan Kaufmann Publishers Inc, San Francisco, CA, USA, 2001: pp. 283–290.

[93] Y. Tian, Y. Zhang, Y. Su, X. Zhang, K.C. Tan, Y. Jin, Balancing Objective Optimization and Constraint Satisfaction in Constrained Evolutionary Multiobjective Optimization, IEEE Trans. Cybern. 52 (2022) 9559–9572. https://doi.org/10.1109/TCYB.2020.3021138.

[94] R. Wang, R.C. Purshouse, P.J. Fleming, Preference-Inspired Coevolutionary Algorithms for Many-Objective Optimization, IEEE Trans. Evol. Computat. 17 (2013) 474–494. https://doi.org/10.1109/TEVC.2012.2204264.

[95] Y. Tian, J. Chen, X. Zhang, Hybrid optimizer combining evolutionary computation and gradient descent for constrained multi-objective optimization, Journal of Computer Applications (in Chinese) 44 (2024) 1386–1392. https://doi.org/https://doi.org/10.11772/j.issn.1001-9081.2023121798.

[96] Qingfu Zhang, Aimin Zhou, Yaochu Jin, RM-MEDA: A Regularity Model-Based Multiobjective Estimation of Distribution Algorithm, IEEE Trans. Evol. Computat. 12 (2008) 41–63. https://doi.org/10.1109/TEVC.2007.894202.

[97] X. Zhang, X. Zheng, R. Cheng, J. Qiu, Y. Jin, A competitive mechanism based multi-objective particle swarm optimizer with fast convergence, Information Sciences 427 (2018) 63–76. https://doi.org/10.1016/j.ins.2017.10.037.

[98] C. He, Y. Tian, Y. Jin, X. Zhang, L. Pan, A radial space division based evolutionary algorithm for many-objective optimization, Applied Soft Computing 61 (2017) 603–621. https://doi.org/10.1016/j.asoc.2017.08.024.

[99] R. Jiao, S. Zeng, C. Li, S. Yang, Y.-S. Ong, Handling Constrained Many-Objective Optimization Problems via Problem Transformation, IEEE Trans. Cybern. 51 (2021) 4834–4847. https://doi.org/10.1109/TCYB.2020.3031642.

[100] R. Cheng, Y. Jin, M. Olhofer, B. Sendhoff, A Reference Vector Guided Evolutionary Algorithm for Many-Objective Optimization, IEEE Trans. Evol. Computat. 20 (2016) 773–791. https://doi.org/10.1109/TEVC.2016.2519378.

[101] Y. Liu, H. Ishibuchi, N. Masuyama, Y. Nojima, Adapting Reference Vectors and Scalarizing Functions by Growing Neural Gas to Handle Irregular Pareto Fronts, IEEE Trans. Evol. Computat. (2020) 1–1. https://doi.org/10.1109/TEVC.2019.2926151.

[102] Q. Liu, Y. Jin, M. Heiderich, T. Rodemann, G. Yu, An Adaptive Reference Vector-Guided Evolutionary Algorithm Using Growing Neural Gas for Many-Objective Optimization of Irregular Problems, IEEE Trans. Cybern. 52 (2022) 2698–2711. https://doi.org/10.1109/TCYB.2020.3020630.

[103] F. Ming, W. Gong, L. Wang, Y. Jin, Constrained Multi-Objective Optimization With Deep Reinforcement Learning Assisted Operator Selection, IEEE/CAA J. Autom. Sinica 11 (2024) 919–931. https://doi.org/10.1109/JAS.2023.123687.

[104] K. Yu, J. Liang, B. Qu, Y. Luo, C. Yue, Dynamic Selection Preference-Assisted Constrained Multiobjective Differential Evolution, IEEE Trans. Syst. Man Cybern, Syst. 52 (2022) 2954–2965. https://doi.org/10.1109/TSMC.2021.3061698.

[105] S. Jiang, S. Yang, A Strength Pareto Evolutionary Algorithm Based on Reference Direction for Multiobjective and Many-Objective Optimization, IEEE Trans. Evol. Computat. 21 (2017) 329–346. https://doi.org/10.1109/TEVC.2016.2592479.

[106] Y. Yuan, H. Xu, B. Wang, B. Zhang, X. Yao, Balancing Convergence and Diversity in Decomposition-Based Many-Objective Optimizers, IEEE Trans. Evol. Computat. 20 (2016) 180–198. https://doi.org/10.1109/TEVC.2015.2443001.

[107] E. Zitzler, M. Laumanns, L. Thiele, SPEA2: Improving the strength pareto evolutionary algorithm, ETH Zurich, 2001. https://doi.org/10.3929/ETHZ-A-004284029.

[108] R. Denysiuk, L. Costa, I. Espírito Santo, Clustering-Based Selection for Evolutionary Many-Objective Optimization, in: T. Bartz-Beielstein, J. Branke, B. Filipič, J. Smith (Eds.), Parallel Problem Solving from Nature – PPSN XIII, Springer International Publishing, Cham, 2014: pp. 538–547. https://doi.org/10.1007/978-3-319-10762-2_53.

[109] G. Liu, Z. Pei, N. Liu, Y. Tian, Subspace segmentation based co-evolutionary algorithm for balancing convergence and diversity in many-objective optimization, Swarm and Evolutionary Computation 83 (2023) 101410. https://doi.org/10.1016/j.swevo.2023.101410.

[110] S. Kukkonen, J. Lampinen, GDE3: The third Evolution Step of Generalized Differential Evolution, in: 2005 IEEE Congress on Evolutionary Computation, IEEE, Edinburgh, Scotland, UK, 2005: pp. 443–450. https://doi.org/10.1109/CEC.2005.1554717.

[111] F. Ming, W. Gong, L. Wang, A Two-Stage Evolutionary Algorithm With Balanced Convergence and Diversity for Many-Objective Optimization, IEEE Trans. Syst. Man Cybern, Syst. 52 (2022) 6222–6234. https://doi.org/10.1109/TSMC.2022.3143657.

[112] E. Zitzler, S. Künzli, Indicator-Based Selection in Multiobjective Search, in: X. Yao, E.K. Burke, J.A. Lozano, J. Smith, J.J. Merelo-Guervós, J.A. Bullinaria, J.E. Rowe, P. Tiňo, A. Kabán, H.-P. Schwefel (Eds.), Parallel Problem Solving from Nature - PPSN VIII, Springer Berlin Heidelberg, Berlin, Heidelberg, 2004: pp. 832–842. https://doi.org/10.1007/978-3-540-30217-9_84.

[113] J. Dong, W. Gong, F. Ming, L. Wang, A two-stage evolutionary algorithm based on three indicators for constrained multi-objective optimization, Expert Systems with Applications 195 (2022) 116499. https://doi.org/10.1016/j.eswa.2022.116499.

[114] J. Yuan, H.-L. Liu, Y.-S. Ong, Z. He, Indicator-Based Evolutionary Algorithm for Solving Constrained Multiobjective Optimization Problems, IEEE Trans. Evol. Computat. 26 (2022) 379–391. https://doi.org/10.1109/TEVC.2021.3089155.

[115] Z.-Z. Liu, Y. Wang, Handling Constrained Multiobjective Optimization Problems With Constraints in Both the Decision and Objective Spaces, IEEE Trans. Evol. Computat. 23 (2019) 870–884. https://doi.org/10.1109/TEVC.2019.2894743.

[116] K. Qiao, J. Liang, K. Yu, C. Yue, H. Lin, D. Zhang, B. Qu, Evolutionary Constrained Multiobjective Optimization: Scalable High-Dimensional Constraint Benchmarks and Algorithm, IEEE Trans. Evol. Computat. 28 (2024) 965–979. https://doi.org/10.1109/TEVC.2023.3281666.

[117] S. Zapotecas Martínez, C.A. Coello Coello, A multi-objective particle swarm optimizer based on decomposition, in: Proceedings of the 13th Annual Conference on Genetic and Evolutionary Computation, ACM, Dublin Ireland, 2011: pp. 69–76. https://doi.org/10.1145/2001576.2001587.

[118] J. Zou, R. Sun, Y. Liu, Y. Hu, S. Yang, J. Zheng, K. Li, A Multipopulation Evolutionary Algorithm Using New Cooperative Mechanism for Solving Multiobjective Problems With Multiconstraint, IEEE Trans. Evol. Computat. 28 (2024) 267–280. https://doi.org/10.1109/TEVC.2023.3260306.

[119] J. Molina, L.V. Santana, A.G. Hernández-Díaz, C.A. Coello Coello, R. Caballero, g-dominance: Reference point based dominance for multiobjective metaheuristics, European Journal of Operational Research 197 (2009) 685–692. https://doi.org/10.1016/j.ejor.2008.07.015.

[120] R. Jiao, B. Xue, M. Zhang, A Multiform Optimization Framework for Constrained Multiobjective Optimization, IEEE Trans. Cybern. 53 (2023) 5165–5177. https://doi.org/10.1109/TCYB.2022.3178132.

[121] H. Chen, Y. Tian, W. Pedrycz, G. Wu, R. Wang, L. Wang, Hyperplane Assisted Evolutionary Algorithm for Many-Objective Optimization Problems, IEEE Trans. Cybern. 50 (2020) 3367–3380. https://doi.org/10.1109/TCYB.2019.2899225.

[122] Q. Lin, J. Li, Z. Du, J. Chen, Z. Ming, A novel multi-objective particle swarm optimization with multiple search strategies, European Journal of Operational Research 247 (2015) 732–744. https://doi.org/10.1016/j.ejor.2015.06.071.

[123] Y. Liu, D. Gong, J. Sun, Y. Jin, A Many-Objective Evolutionary Algorithm Using A One-by-One Selection Strategy, IEEE Trans. Cybern. 47 (2017) 2689–2702. https://doi.org/10.1109/TCYB.2016.2638902.

[124] J. Jiang, J. Wu, J. Luo, X. Yang, Z. Huang, MOBCA: Multi-Objective Besiege and Conquer Algorithm, Biomimetics 9 (2024) 316. https://doi.org/10.3390/biomimetics9060316.

[125] Y. Yuan, H. Xu, B. Wang, X. Yao, A New Dominance Relation-Based Evolutionary Algorithm for Many-Objective Optimization, IEEE Trans. Evol. Computat. 20 (2016) 16–37. https://doi.org/10.1109/TEVC.2015.2420112.

[126] H. Li, Q. Zhang, J. Deng, Biased Multiobjective Optimization and Decomposition Algorithm, IEEE Trans. Cybern. 47 (2017) 52–66. https://doi.org/10.1109/TCYB.2015.2507366.

[127] F. Ming, W. Gong, L. Wang, L. Gao, A Constraint-Handling Technique for Decomposition-Based Constrained Many-Objective Evolutionary Algorithms, IEEE Trans. Syst. Man Cybern, Syst. 53 (2023) 7783–7793. https://doi.org/10.1109/TSMC.2023.3299570.

[128] C.R. Raquel, P.C. Naval, An effective use of crowding distance in multiobjective particle swarm optimization, in: Proceedings of the 7th Annual Conference on Genetic and Evolutionary Computation, ACM, Washington DC USA, 2005: pp. 257–264. https://doi.org/10.1145/1068009.1068047.

[129] E. Zitzler, L. Thiele, M. Laumanns, C.M. Fonseca, V.G. Da Fonseca, Performance assessment of multiobjective optimizers: an analysis and review, IEEE Trans. Evol. Computat. 7 (2003) 117–132. https://doi.org/10.1109/TEVC.2003.810758.
[130] T.H.W. Bäck, A.V. Kononova, B. Van Stein, H. Wang, K.A. Antonov, R.T. Kalkreuth, J. De Nobel, D. Vermetten, R. De Winter, F. Ye, Evolutionary Algorithms for Parameter Optimization—Thirty Years Later, Evolutionary Computation 31 (2023) 81–122. https://doi.org/10.1162/evco_a_00325.
[131] A. LaTorre, D. Molina, E. Osaba, J. Poyatos, J. Del Ser, F. Herrera, A prescription of methodological guidelines for comparing bio-inspired optimization algorithms, Swarm and Evolutionary Computation 67 (2021) 100973. https://doi.org/10.1016/j.swevo.2021.100973.
[132] S. Holm, A Simple Sequentially Rejective Multiple Test Procedure, Scandinavian Journal of Statistics 6 (1979) 65–70.
[133] M. Aickin, H. Gensler, Adjusting for multiple testing when reporting research results: the Bonferroni vs Holm methods., Am J Public Health 86 (1996) 726–728. https://doi.org/10.2105/AJPH.86.5.726.